\documentclass[pdflatex,sn-mathphys-num]{sn-jnl}

\usepackage{graphicx}%
\usepackage{multirow}%
\usepackage{amsmath,amssymb,amsfonts}%
\usepackage{amsthm}%
\usepackage{mathrsfs}%
\usepackage[title]{appendix}%
\usepackage{xcolor}%
\usepackage{textcomp}%
\usepackage{manyfoot}%
\usepackage{booktabs}%
\usepackage{algorithm}%
\usepackage{algorithmicx}%
\usepackage{algpseudocode}%
\usepackage{listings}%

\usepackage{tikz}           
\usepackage{caption}        
\usepackage{subcaption}     

\usetikzlibrary{positioning}

\theoremstyle{thmstyleone}%
\theoremstyle{thmstyletwo}%

\theoremstyle{thmstylethree}%

\begin{document}

\title[Article Title]{Modeling Psychological Profiles in Volleyball via Mixed-Type Bayesian Networks}


\author[1]{\fnm{Maria} \sur{Iannario}}\email{maria.iannario@unina.it}

\author[2]{\fnm{Dae-Jin} \sur{Lee}}\email{dae-jin.lee@ie.edu}

\author*[2]{\fnm{Manuele} \sur{Leonelli}}\email{manuele.leonelli@ie.edu}

\affil*[1]{\orgdiv{Department of Political Sciences}, \orgname{University of Naples Federico II}, \orgaddress{\street{Via Rodinò 22}, \city{Napoli}, \postcode{80133}, \country{Italy}}}

\affil[2]{\orgdiv{School of Science and Technology}, \orgname{IE University}, \orgaddress{\street{ Paseo de la Castellana 259}, \city{Madrid}, \postcode{28046}, \country{Spain}}}


\abstract{Psychological attributes rarely operate in isolation: coaches reason about networks of related traits. We analyze a new dataset of 164 female volleyball players from Italy’s C and D leagues that combines standardized psychological profiling with background information. To learn directed relationships among mixed-type variables (ordinal questionnaire scores, categorical demographics, continuous indicators), we introduce latent MMHC, a hybrid structure learner that couples a latent Gaussian copula and a constraint-based skeleton with a constrained score-based refinement to return a single DAG. We also study a bootstrap-aggregated variant for stability. In simulations spanning sample size, sparsity, and dimension, latent Max-Min Hill-Climbing (MMHC) attains lower structural Hamming distance and higher edge recall than recent copula-based learners while maintaining high specificity. Applied to volleyball, the learned network organizes mental skills around goal setting and self-confidence, with emotional arousal linking motivation and anxiety, and locates Big-Five traits (notably neuroticism and extraversion) upstream of skill clusters. Scenario analyses quantify how improvements in specific skills propagate through the network to shift preparation, confidence, and self-esteem. The approach provides an interpretable, data-driven framework for profiling psychological traits in sport and for decision support in athlete development.}




\maketitle

\section{Introduction}\label{sec1}

Elite and amateur sport increasingly recognize that psychological factors are central to performance. Meta-analytic reviews show that key psychological constructs (such as motivation, self-efficacy, affective states, and trained mental skills) are systematically related to competitive outcomes \citep{ayranci2025complex,brown2017effects,lochbaum2021profile}. Personality traits and self-esteem further shape how athletes respond to pressure and to coaching interventions \citep{allen2013personality,allen2014role,laborde2020trait}. In parallel, the routine collection of standardized psychological questionnaires alongside demographic, training, and performance records has expanded, creating opportunities for data-driven modeling of mental skills in sport \citep{albert2017handbook,mcgarry2009applied}.

Beyond single-variable analyses, coaches and practitioners typically reason about networks of interrelated traits and contexts \citep{laborde2020trait}. Recent work across both team and individual sports adopts an explicitly multivariate perspective that links personality, coping and mental skills, and performance, and shows that effects frequently operate indirectly through intermediate psychological processes \citep{fabbricatore2023component,kalinowski2020mediational,iannario2023dyadic}. Understanding how psychological attributes influence one another, and how they collectively relate to performance, requires models that represent multivariate dependence and conditional structure.

Bayesian networks (BNs) provide such a framework: they encode conditional dependencies in a directed graph and support probabilistic reasoning under uncertainty \citep{koller2009probabilistic,lauritzen1996graphical}. In psychology and psychopathology, BNs are now widely used to assess interrelations among traits and symptoms and to probe admissible causal structures from observational data \citep{briganti2023tutorial,briganti2024using}. In sport, however, particularly for modeling athletes’ psychological profiles, BN applications remain rather limited \citep{fuster2014bayesian,fuster2015team,ponseti2019self}. More broadly across sports analytics, BN methods are still underrepresented relative to regression, SEM, and black-box machine-learning approaches, despite successful demonstrations in outcome prediction, performance assessment, injury prognosis, and tactical decision modeling \citep{constantinou2019dolores,constantinou2012pi,d2023bayesian,yung2025using,zhou2024probability}.

Sport psychology datasets typically combine Likert-scale questionnaire scores (ordinal), demographic or role information (categorical), and training or performance indicators (often continuous) \citep{fabbricatore2023component,iannario2023dyadic}. In much applied BN work, this heterogeneity is handled by discretizing continuous variables or collapsing categories before analysis. However, ad hoc binning can distort dependence structure, reduce statistical power, and yield graphs that change with the chosen cut points \citep{beuzen2018comparison,nojavan2017comparative}. These limitations have been noted in recent performance-analytics applications as well \citep{d2023bayesian}. Accordingly, approaches that avoid ad hoc discretization are desirable.

There is growing interest in BNs for mixed‐type variables, a setting historically less developed than purely discrete or Gaussian formulations. Such data are precisely the focus of the present work. Classical conditional linear Gaussian models allow discrete and continuous variables but impose parent–child restrictions that limit expressiveness \citep{lauritzen1989graphical}. More recent work leverages latent Gaussian copulas with rank-based estimation to model mixed marginals while retaining tractable structure learning \citep{hoff2007extending,cui2016copula,cai2022causal,castelletti2024learning}. Most proposals to date are \emph{constraint-based}, recovering the graph through sequences of conditional-independence tests that add, remove, and orient edges.

Building on these developments, we adopt the “hybrid” strategy familiar from discrete and Gaussian BNs, where a constraint-based phase is followed by a score-based refinement whose search space is restricted by the first phase (e.g., Max–Min Hill-Climbing; \citealp{tsamardinos2006max}). To our knowledge, we are the first to bring this approach to mixed data under a latent Gaussian copula: we extend latent–copula learners \citep{cui2016copula,cai2022causal} with a constrained score-based refinement. We term the resulting algorithm \emph{latent MMHC}. Unlike purely constraint-based copula learners, which may leave some edge directions unresolved, latent MMHC returns a single oriented Directed Acyclic Graph (DAG). We also study a bootstrap-aggregated “stable” variant that repeats the procedure on resamples and aggregates edge-selection frequencies to produce a consensus DAG \citep{scutari2013identifying,caravagna2021learning}. 

In simulations spanning variations in sample size, sparsity, and dimensionality, latent MMHC demonstrates improved structural recovery and edge recall compared to competing methods, while maintaining high specificity. We also discuss the associated computational trade-offs and provide open-source \texttt{R} code to facilitate reproducibility.
Motivating our methodological development is a dataset on \emph{female volleyball players} from Italy’s C and D leagues in the Campania region, which combines psychological questionnaires with demographic and experience information. We apply our \emph{latent MMHC} algorithm and, for robustness, a bootstrap-aggregated \emph{stable} version to learn an interpretable directed network from these mixed-type variables. The resulting model offers a transparent view of how psychological attributes are conditionally related and provides a basis for broader analyses: identifying traits that may influence others, highlighting potential connectors, and running scenario exercises to explore how improvements in one skill can reverberate through the system. In this way, routine profiling data are transformed into insights that can inform training and athlete development.

The paper is structured as follows. Section~\ref{sec:method} presents the mixed-type BN framework and introduces the latent MMHC algorithm and its stable variant. Section~\ref{sec:sim} benchmarks the proposed methods across sample sizes, sparsity levels, and dimensions. Section~\ref{sec:volley} applies the approach to the volleyball dataset to derive interpretable, coaching-relevant insights. Section~\ref{sec:conc} summarizes the findings and outlines avenues for future work.

\section{Methodology}
\label{sec:method}
\subsection{Graphical Models and Directed Acyclic Graphs}

Let $\mathbf{X} = (X_1, \dots, X_p)$ be a vector of random variables.  A \emph{directed acyclic graph} (DAG) $\mathcal{G} = (V, E)$ consists of a set of nodes $V = \{1, \dots, p\}$ and a set of directed edges $E \subset V \times V$ with no directed cycles. Each node $j \in V$ corresponds to a random variable $X_j$, and a directed edge $(i \rightarrow j) \in E$ indicates that $X_i$ is a direct parent of $X_j$.

The DAG encodes a factorization of the joint distribution $P$ of $\mathbf{X}$ according to the local Markov property \citep{koller2009probabilistic,lauritzen1996graphical}:
\begin{equation}
\label{eq:factorization}
P(\mathbf{X}) = \prod_{j=1}^p P(X_j \mid \mathbf{X}_{\text{pa}(j)}),
\end{equation}
where $\text{pa}(j)$ denotes the set of indices corresponding to the parents of node $j$ in $\mathcal{G}$. This factorization implies a collection of conditional independence relations, which are entailed by the structure of the graph. By breaking down the joint distribution into local components, the DAG provides a compact and interpretable representation of potentially high-dimensional dependencies.

The set of conditional independencies encoded by the DAG can be formally characterized using the concept of \emph{d-separation} \citep{pearl2009causality}. Let $\pi$ be a path between nodes $i$ and $j$ in the graph. The path $\pi$ is said to be \emph{blocked} by a set of nodes $S \subset V$ if there exists a node $k$ on the path such that one of the following conditions holds:
\begin{enumerate}
  \item The path passes through a chain or fork structure $X_u \rightarrow X_k \rightarrow X_v$ or $X_u \leftarrow X_k \rightarrow X_v$, and $X_k \in S$.
  \item The path passes through a collider structure $X_u \rightarrow X_k \leftarrow X_v$, and neither $X_k$ nor any of its descendants belong to $S$.
\end{enumerate}
If all paths between $X_i$ and $X_j$ are blocked by $S$, we say that $X_i$ and $X_j$ are \emph{d-separated} given $S$ in $\mathcal{G}$, and denote this by $X_i \perp\!\!\!\perp X_j \mid \mathbf{X}_S [\mathcal{G}]$. The global Markov property, stating that all d-separation statements in the graph correspond to conditional independencies in the distribution, is equivalent to the local Markov property expressed in Eq.~(\ref{eq:factorization}) under standard assumptions \citep{lauritzen1996graphical,koller2009probabilistic}. These two characterizations offer distinct but equivalent perspectives on the dependency structure encoded by the DAG.

\begin{figure}
\centering

\begin{subfigure}[b]{0.22\textwidth}
\centering
\begin{tikzpicture}[->,>=latex, node distance=0.4cm,font=\scriptsize]
\node[draw, circle] (X1) {$1$};
\node[draw, circle] (X3) [right=of X1] {$3$};
\node[draw, circle] (X2) [right=of X3] {$2$};

\draw[->] (X1) -- (X3);
\draw[->] (X2) -- (X3);
\end{tikzpicture}
\caption{}
\end{subfigure}
\hfill
\begin{subfigure}[b]{0.22\textwidth}
\centering
\begin{tikzpicture}[->,>=latex, node distance=0.4cm,font=\scriptsize]
\node[draw, circle] (X1) {$1$};
\node[draw, circle] (X3) [right=of X1] {$3$};
\node[draw, circle] (X2) [right=of X3] {$2$};

\draw[->] (X1) -- (X3);
\draw[->] (X3) -- (X2);
\end{tikzpicture}
\caption{}
\end{subfigure}
\hfill
\begin{subfigure}[b]{0.22\textwidth}
\centering
\begin{tikzpicture}[->,>=latex, node distance=0.4cm,font=\scriptsize]
\node[draw, circle] (X1) {$1$};
\node[draw, circle] (X3) [right=of X1] {$3$};
\node[draw, circle] (X2) [right=of X3] {$2$};

\draw[->] (X3) -- (X1);
\draw[->] (X3) -- (X2);
\end{tikzpicture}
\caption{}
\end{subfigure}
\hfill
\begin{subfigure}[b]{0.22\textwidth}
\centering
\begin{tikzpicture}[-,>=latex, node distance=0.4cm,font=\scriptsize]
\node[draw, circle] (X1) {$1$};
\node[draw, circle] (X3) [right=of X1] {$3$};
\node[draw, circle] (X2) [right=of X3] {$2$};

\draw[-] (X1) -- (X3);
\draw[-] (X2) -- (X3);
\end{tikzpicture}
\caption{}
\end{subfigure}

\caption{Illustration of skeletons, colliders, and Markov equivalence. DAGs (a), (b), and (c) share the same skeleton $1 - 3 - 2$, but only DAG (a) contains the collider $1 \rightarrow 3 \leftarrow 2$. DAGs (b) and (c) are Markov equivalent and belong to the same equivalence class, represented by the CPDAG in panel (d), where undirected edges reflect ambiguous orientations.}
\label{fig:cpdag-example}
\end{figure}
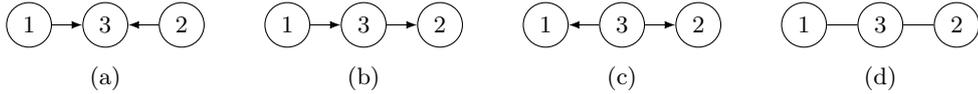

In practice, several different DAGs can encode the same set of conditional independencies. This motivates the notion of \emph{Markov equivalence}: two DAGs are said to be Markov equivalent if they entail the same set of d-separation statements \citep{verma1990equivalence}. A fundamental result establishes that two DAGs are Markov equivalent if and only if they have the same skeleton (i.e., the undirected graph obtained by removing edge directions) and the same set of colliders. The collection of all DAGs that are Markov equivalent forms a \emph{Markov equivalence class}, which can be uniquely represented by a \emph{completed partially directed acyclic graph} (CPDAG) \citep{andersson1997characterization}. A CPDAG is a mixed graph in which an edge is directed if its orientation is shared across all DAGs in the equivalence class, and undirected if at least one DAG assigns each possible orientation. Fig.~\ref{fig:cpdag-example} illustrates these concepts.

\subsection{Learning Graphical Models from Data}

Learning a DAG is a central task in probabilistic modeling when the dependence structure among a collection of variables is unknown and must be inferred from data. Given a dataset $\mathcal{D} = \{\mathbf{x}^{(1)}, \dots, \mathbf{x}^{(n)}\}$ consisting of $n$ independent observations of a random vector $\mathbf{X}$, the goal is to recover a DAG $\mathcal{G}$ whose structure encodes a set of conditional independence relations that are compatible with the joint distribution of $\mathbf{X}$. In practice, this means finding a graph such that the conditional independencies implied by the DAG closely match those supported by the observed data.

Three broad classes of algorithms have been developed to tackle this problem: constraint-based, score-based, and hybrid methods \citep{kitson2023survey,scanagatta2019survey}.

Constraint-based methods rely on performing a series of conditional independence tests among the observed variables \citep{spirtes2000causation}. These methods use the principle that conditional independencies in the distribution can be mapped to d-separation statements in a DAG under appropriate modeling assumptions. A prototypical example is the PC algorithm \citep{colombo2014order}, which starts from a fully connected undirected graph and iteratively removes edges between variables found to be conditionally independent given some subset of the remaining variables. Once the skeleton is estimated, a set of orientation rules is applied to identify colliders and propagate edge directions without introducing cycles or new colliders. The output is typically a CPDAG representing the Markov equivalence class of DAGs consistent with the observed conditional independencies.

Score-based methods, in contrast, treat the problem of structure learning as an optimization task. A scoring function is defined over the space of DAGs, quantifying how well each graph explains the data according to a chosen statistical criterion, such as the Bayesian Information Criterion (BIC) \citep{Schwarz1978}. The graph with the highest score is typically sought using a greedy or heuristic search strategy (e.g. tabu search), where each step considers local modifications to the current graph by adding, deleting, or reversing a single edge while maintaining acyclicity \citep{scutari2010learning}. These methods always return a single DAG that maximizes the chosen score over the explored search space.

Hybrid approaches combine features from both constraint-based and score-based methods. One widely used example is the Max-Min Hill-Climbing (MMHC) algorithm \citep{tsamardinos2006max}. In its first phase, a constraint-based procedure is used to estimate the skeleton of the graph. In the second phase, a score-based search is performed, where edge additions, deletions, or reversals are only allowed if they are consistent with the previously identified skeleton. This is typically enforced by defining a \emph{blacklist} that prohibits edges not present in the skeleton from being considered during the search.

No single approach consistently outperforms the others; their relative performance depends on factors such as sample size, dimensionality, and distributional properties of the data \citep{constantinou2021large,scutari2019learns}.

\subsection{Structure Learning in the Gaussian Case}

Gaussian BNs \citep{geiger1994learning} provide a natural and analytically tractable setting for structure learning, forming the basis for methods applicable to more general or mixed data types. Under this model, each variable is expressed as a linear function of its parents in the DAG, plus an independent Gaussian noise term. That is, for each node $j$ in the graph,
\begin{equation}
X_j = \sum_{k \in \text{pa}(j)} \beta_{jk} X_k + \varepsilon_j, \quad \varepsilon_j \sim \mathcal{N}(0, \sigma_j^2),
\end{equation}
where the residuals $(\varepsilon_1, \dots, \varepsilon_p)$ are mutually independent. This formulation defines a structural equation model (SEM), in which the regression coefficients $\beta_{jk}$ encode the edge structure of the DAG.

Let $L$ be a $p \times p$ lower triangular matrix with unit diagonal, such that $L_{uv} \neq 0$ if and only if $u \rightarrow v$ in $\mathcal{G}$, and let $D$ be a diagonal matrix with entries $\sigma_j^2$. The SEM system can then be compactly written as  $L^\top \mathbf{X} = \boldsymbol{\varepsilon}, \quad \boldsymbol{\varepsilon} \sim \mathcal{N}_p(\mathbf{0}, D).$ It follows that the joint distribution of $\mathbf{X}$ is multivariate normal, with covariance matrix $\Sigma = L^{-\top} D L^{-1}$ and precision matrix $\Omega = \Sigma^{-1} = L D^{-1} L^\top$ \citep{richardson2002ancestral}. This decomposition makes explicit how the DAG structure encoded in $L$ determines the joint distribution, and is particularly useful for efficient score evaluation in structure learning.

In score-based learning of Gaussian BNs, the structure is inferred by optimizing a likelihood-based criterion. To simplify notation, we assume without loss of generality that the data are standardized, with $\mathbb{E}[X_j] = 0$ and $\mathrm{Var}(X_j) = 1$ for all $j$. Under this setup, and adopting a SEM perspective, the log-likelihood can be decomposed into a sum of nodewise contributions \citep{scutari2010learning}:
\begin{equation}
\label{eq:sem}
\ell(\mathcal{G}) = -\frac{n}{2} \sum_{j=1}^p \log(\widehat{\sigma}_j^2),
\end{equation}
where $\widehat{\sigma}_j^2$ is the residual variance from the regression of $X_j$ on its parents. An alternative but equivalent expression based on the coefficient of determination $R_j^2$ is given by
\begin{equation}
\label{eq:gauss}
\ell(\mathcal{G}) = \text{const} - \frac{n}{2} \sum_{j=1}^p \log(1 - R_j^2),
\end{equation}
which follows from standard linear regression theory under Gaussian errors \citep{murphy2012machine}.

In constraint-based approaches, structure learning proceeds by testing conditional independencies implied by the Gaussian distribution. Under multivariate normality, the conditional independence $X_i \perp\!\!\!\perp X_j \mid \mathbf{X}_S$ is equivalent to $\rho_{ij \cdot S} = 0$, where $\rho_{ij \cdot S}$ denotes the partial correlation between $X_i$ and $X_j$ given $S$. This allows conditional independence testing to be performed using standard tools from Gaussian theory, based on estimates of partial correlations and their sampling distribution \citep{kalisch2007estimating}. This testing procedure forms the basis of the Gaussian version of the PC algorithm and related constraint-based methods.

\subsection{Mixed Data and Latent Gaussian Copula Models}

In many real-world applications, datasets consist of a vector of observed variables $\mathbf{X} = (X_1, \dots, X_p)$, which may include a combination of continuous, ordinal, binary, and count components. While BNs are widely used to model such dependencies, structure learning from mixed data remains comparatively underexplored. The most common approach is the \emph{conditional linear Gaussian} model \cite{lauritzen1989graphical}, which assumes Gaussian linear regressions for continuous variables and multinomial distributions for discrete ones. However, it imposes restrictive constraints on the structure of the graph, notably disallowing continuous parents of discrete nodes, and can struggle to represent complex dependencies.

To address these limitations, recent approaches model mixed data using \emph{latent Gaussian copulas} \cite{cai2022causal,castelletti2024learning,cui2016copula,harris2013pc}, where each observed variable is assumed to be either a monotonic transformation (for continuous data) or a discretization (for ordinal or binary data) of a standard normal latent variable. Specifically, for each observed variable $X_j$, we assume the existence of a latent variable $Z_j \sim \mathcal{N}(0, 1)$ such that
\[
Z_j = \Phi^{-1}(F_j(X_j)),
\]
where $F_j$ is the marginal cumulative distribution function of $X_j$, and $\Phi^{-1}$ is the quantile function of the standard normal. The vector $\mathbf{Z} = (Z_1, \dots, Z_p)$ is then modeled as a multivariate Gaussian:
\[
\mathbf{Z} \sim \mathcal{N}_p(0, \Sigma),
\]
where $\Sigma$ is a correlation matrix that encodes the dependence structure among the latent variables. This construction induces a Gaussian copula over the observed variables $\mathbf{X}$, while allowing arbitrary marginal distributions for each component. Alternative approaches not relying on a latent Gaussian layer, such as ordinal regression, conditional likelihoods, or structured local models, have also been explored in the literature \cite{andrews2018scoring,raghu2018evaluation,talvitie2019learning,tsagris2018constraint}.

The key idea in latent Gaussian copula models is that the dependence structure among the observed variables $\mathbf{X}$ is governed by the latent correlation matrix $\Sigma$ defining the joint distribution of the latent variables $\mathbf{Z}$. When marginals are continuous, conditional independencies in $\mathbf{Z}$ carry over to $\mathbf{X}$ \citep{liu2009nonparanormal}; with discrete margins, additional dependencies may arise but are typically viewed as secondary \citep{dobra2011copula}.

One strategy to estimate $\Sigma$ is based on Hoff’s extended rank likelihood \citep{hoff2007extending}, which avoids explicit modeling of the marginal distributions by treating the observed ranks as constraints on the latent variables. This approach underlies the \emph{copula PC} method \citep{cui2016copula}, where the correlation matrix is estimated via Gibbs sampling. A fully Bayesian extension has also been proposed \citep{castelletti2024learning}, combining the rank likelihood with a DAG-Wishart prior. A known challenge in this context is the effective loss of sample size when estimating partial correlations involving sparse levels.

An alternative strategy, used in the \emph{latent PC} method \citep{cai2022causal}, estimates the latent correlation matrix directly from the observed data using pairwise Kendall’s $\tau$ coefficients. For each pair of variables $(X_j, X_k)$, the latent correlation $\sigma_{jk}$ is estimated via the identity $\sigma_{jk} = \sin\left(\pi\tau_{jk}/2 \right)$, where $\tau_{jk}$ denotes the empirical Kendall correlation between $X_j$ and $X_k$. This provides a closed-form, rank-based estimator of $\Sigma$ that is simple to compute and robust to marginal distributional assumptions.

Irrespective of the method used to estimate the latent correlation matrix $\Sigma$, structure learning proceeds by testing conditional independencies via partial correlations computed from $\Sigma$. These tests form the basis of constraint-based algorithms such as the PC algorithm, applied directly to the latent space. By leveraging the Gaussian properties of the latent variables, standard methods for continuous data can be extended to mixed settings without requiring parametric assumptions on the observed marginals.

\subsection{Hybrid Structure Learning from Latent Gaussian Copulas}

Most existing structure learning approaches based on latent Gaussian copulas rely on constraint-based algorithms such as the PC algorithm \citep{cui2016copula,cai2022causal}. While these methods can consistently recover the Markov equivalence class under standard assumptions, they do not return a unique DAG. The only exception is the fully Bayesian approach of \cite{castelletti2024learning}, which samples from the posterior distribution over DAGs. However, this method is computationally demanding and may become impractical as the number of variables increases.

We introduce a new class of hybrid algorithms for structure learning from mixed data, which we call \emph{latent MMHC}. These methods extend existing PC-based copula approaches by incorporating a score-based refinement phase, following the MMHC framework \citep{tsamardinos2006max}. The graph skeleton is first estimated via the \emph{latent PC} algorithm, using Kendall’s $\tau$ to estimate the latent correlation matrix. A constrained score-based search is then used to orient edges and return a DAG. To our knowledge, this is the first hybrid structure learning approach for mixed data based on latent Gaussian copulas, addressing a key gap in the literature: existing latent copula methods rely almost exclusively on constraint-based techniques, which return only equivalence classes and may struggle to orient edges reliably.

In the refinement step, we define a \emph{blacklist} that prohibits any edge not present in the skeleton obtained during the constraint-based phase. We then apply a tabu search over the space of DAGs, scoring candidates using either the SEM-based likelihood in Eq.~(\ref{eq:sem}), computed from rank-based latent variables, or a Gaussian log-likelihood approximation based on the latent correlation matrix in Eq.~(\ref{eq:gauss}). These two variants are referred to as latent MMHC (SEM) and latent MMHC (Gauss), respectively. 

We focus on latent MMHC because prior work has shown the latent PC method to outperform copula PC in both accuracy and scalability \citep{cai2022causal}, particularly when dealing with high-dimensional or sparse data. However, the same hybrid strategy could be applied using copula PC in the first phase, resulting in a copula MMHC variant. 

Although both copula PC and latent PC are consistent under appropriate conditions, our hybrid procedure does not inherit these guarantees. As discussed in \cite{tsamardinos2006max}, the score-based refinement phase in MMHC lacks theoretical consistency guarantees, since it relies on greedy local search which may converge to suboptimal DAGs even in the infinite-sample limit. Our method shares this limitation. Nevertheless, hybrid algorithms like MMHC have been shown to perform competitively in practice, often improving edge orientation and reducing structural errors compared to purely constraint-based methods. In this spirit, our approach seeks to combine the robustness of latent copula estimators with the practical strengths of score-based refinement.

All proposed methods are implemented in \texttt{R}, using the \texttt{pcalg} package \citep{kalisch2012causal} for skeleton estimation via the PC algorithm and the \texttt{bnlearn} package \citep{scutari2010learning} for score-based search, and are freely available at \url{https://github.com/manueleleonelli/latent\_mmhc}. Both libraries support flexible, high-dimensional structure learning, and \texttt{bnlearn} allows custom scoring functions, enabling direct implementation of the SEM-based and Gaussian log-likelihood scores. Since the final output is a \texttt{bnlearn} DAG object, it integrates seamlessly with the package’s tools for probabilistic inference, model comparison, visualization, and bootstrapping, and is fully compatible with the broader ecosystem of bnlearn-based packages for extended analysis \citep{leonelli2023sensitivity}.

To further improve the stability of the learned structures, we implement a bootstrap-based aggregation scheme that we refer to as \emph{stable latent MMHC}. This procedure is based on the widely used nonparametric bootstrap for BNs \citep{caravagna2021learning,scutari2013identifying}, in which the structure learning algorithm is applied to multiple bootstrap replicates of the data. The resulting DAGs are aggregated into a consensus structure using the \texttt{bnlearn} functions \texttt{averaged.network} and \texttt{inclusion.threshold}, which compute edge selection frequencies and produce a final graph containing only those edges that appear with sufficient support across replicates. This extension enhances the robustness of our method to small-sample variability and complements the underlying statistical strength of the latent copula framework.

\section{Simulation Study}
\label{sec:sim}
\subsection{Data-Generating Mechanism}

We simulate data from a Gaussian copula model with an underlying DAG and mixed-type marginals. The procedure follows \cite{castelletti2024learning,cui2016copula}, and is designed to reflect realistic features observed in applied problems, such as our volleyball dataset.

Given integers $n$ (sample size), $p$ (number of variables), and $s$ (edge sparsity parameter), data are generated as follows:

\begin{enumerate}
\item \textbf{DAG structure:} We construct a weighted adjacency matrix $A \in \mathbb{R}^{p \times p}$ corresponding to a DAG where edges are only allowed from lower-indexed to higher-indexed nodes (i.e., $A_{ij} \neq 0$ only if $j < i$). This guarantees acyclicity by design, as the resulting graph respects a topological ordering.  For each pair $(i, j)$ with $i > j$, an edge from $j$ to $i$ is included independently with probability $s / (p - 1)$, and the corresponding entry $A_{ij}$ is assigned a random weight in $[0.1, 1]$ with a random sign. This procedure yields a sparse DAG in which the expected number of neighbors per node is $s$.

  \item \textbf{Latent Gaussian data:} Latent variables $\mathbf{Z} \in \mathbb{R}^{n \times p}$ are generated by sampling $\boldsymbol{\varepsilon} \sim \mathcal{N}_p(0, I)$ and solving the structural equation model $ \mathbf{Z} = \boldsymbol{\varepsilon} (\mathbf{I} - A)^{-T}.$   This yields samples from a multivariate Gaussian distribution faithful to the generated DAG.

  \item \textbf{Observed mixed data:} The first $p/2$ variables are transformed into discrete observations using monotonic transformations of the corresponding latent variables:
  \begin{itemize}
    \item With probability $0.5$, a variable is made \emph{binary} by thresholding the standard normal CDF $\Phi(Z_j)$ at $1 - \theta$ with $\theta \sim \text{Uniform}(0.2, 0.8)$.
    \item Otherwise, it is made \emph{ordinal} by applying the quantile function of a Binomial(4, $\theta$) distribution to $\Phi(Z_j)$, again with $\theta \sim \text{Uniform}(0.2, 0.8)$.
  \end{itemize}
  This construction ensures that all discrete variables are \emph{upstream} of the continuous ones in the DAG, mirroring the structure of our volleyball dataset, and the resulting categories are \emph{unbalanced}, reflecting the unequal frequency distributions commonly seen in real-world data. The remaining $p/2$ variables are kept continuous and correspond directly to the corresponding columns in $\mathbf{Z}$.

\end{enumerate}

\subsection{Experimental Setup and Evaluation Metrics}

We conducted a small simulation study to evaluate the performance of structure learning algorithms for mixed-type data. For each combination of number of variables $p \in \{10, 30\}$, sample size $n \in \{500, 1000, 2500, 5000\}$, and sparsity level $s \in \{2, 5\}$, we generated 100 random DAGs. For each DAG, a dataset of size $n$ was simulated independently.

We considered a suite of structure learning methods, starting with the established \emph{copula PC} and \emph{latent PC} algorithms, with implementation given in \cite{cui2016copula} and \cite{cai2022causal}, respectively. Following suggestions from both papers, we fixed the critical level for conditional independence testing to $\alpha = 0.01$. We then evaluated our proposed \emph{latent MMHC} algorithms under two scoring schemes, using the SEM-based and Gaussian log-likelihoods from Eqs.~(\ref{eq:sem}) and (\ref{eq:gauss}). To assess the impact of a more permissive constraint phase, we also considered versions with a higher significance level $\alpha = 0.05$ in the initial PC step.

\begin{table}
\centering
\caption{Median SHD across simulation settings varying sample size, sparsity, and number of variables; bold highlights the best-performing method in each configuration.}
\label{tab:shd}
\footnotesize
\renewcommand{\arraystretch}{1.05}
\begin{tabular}{lcccccccc}
\toprule
\textbf{Method} & \multicolumn{4}{c}{\textbf{s = 2}} & \multicolumn{4}{c}{\textbf{s = 5}} \\
\cmidrule(lr){2-5} \cmidrule(lr){6-9}
& 500 & 1000 & 2500 & 5000 & 500 & 1000 & 2500 & 5000 \\
\midrule
\multicolumn{9}{l}{\textbf{p = 10}} \\
copula PC                & 7   & 7   & 6   & 5   & 22  & 22  & 20  & 21  \\
latent PC          & 7   & 6   & 5   & 5.5 & 21  & 22  & 20  & \textbf{19} \\
copula score-based      & 7   & 7   & 8   & 8   & 25  & 24  & 26  & 26  \\
SEM score-based          & 7   & 7   & 5.5 & 5   & 23  & 22  & 23  & 23  \\
latent score-based          & 13  & 12  & 9   & 9   & 27.5& 28  & 27  & 27  \\
latent MMHC (Gauss, 0.01) & 6   & 6   & 4   & 5   & \textbf{20} & \textbf{21} & \textbf{19} & 19.5 \\
latent MMHC (Gauss, 0.05)& 7   & 7   & 5   & 6   & 20.5 & \textbf{21} & \textbf{19} & 20 \\
latent MMHC (SEM, 0.01) & \textbf{5}   & 4.5 & \textbf{3}   & \textbf{2}   & 21  & \textbf{21}  & 20  & 19.5 \\
latent MMHC (SEM, 0.05) & \textbf{5}   & \textbf{4}   & \textbf{3}   & 2.5 & 21  & \textbf{21}  & 20  & \textbf{19} \\
\addlinespace
\midrule
\multicolumn{9}{l}{\textbf{p = 30}} \\
copula PC               & 24  & 22  & 21  & 18.5 & 64.5& 61  & 58  & 56.5 \\
latent PC        & 30  & 30  & 27  & 25   & 60  & 53  & 48  & 44.5 \\
copula score-based      & 33  & 28  & 26  & 27   & 91  & 90  & 93  & 103  \\
SEM score-based          & 30  & 25  & 23  & 23   & 72  & 74.5& 87  & 100 \\
latent score-based         & 135.5 & 112 & 91  & 74.5 & 233 & 216.5 & 208 & 200.5 \\
latent MMHC (Gauss, 0.01) & 29  & 29  & 25  & 22   & 59  & \textbf{48}  & 45  & 43 \\
latent MMHC (Gauss, 0.05) & 37  & 36  & 32.5& 31   & 58  & 50  & 46  & 45 \\
latent MMHC (SEM, 0.01) & \textbf{17}  & \textbf{12}  & \textbf{8}  & \textbf{8}   & \textbf{55}  & 49  & \textbf{44} & \textbf{41} \\
latent MMHC (SEM, 0.05) & 18  & 13  & 9   & 9     & 56  & \textbf{48}  & 46  & 42.5 \\
\bottomrule
\end{tabular}

\end{table}

Lastly, for comprehensiveness, we included three score-based methods inspired by our framework. The \emph{copula score-based} method uses the Gaussian log-likelihood with correlation matrices estimated via Gibbs sampling; the \emph{latent score-based} method uses Kendall’s $\tau$ for latent correlation estimation; and the \emph{SEM score-based} method uses the SEM formulation directly on rank-transformed data. All score-based methods were implemented using tabu search for structure optimization.

We evaluated each method using four performance metrics. The \emph{Structural Hamming Distance} (SHD) counts the number of edge insertions, deletions, and reversals required to transform the estimated DAG into the true DAG; lower SHD values indicate better structural accuracy \cite{tsamardinos2006max}. \emph{Sensitivity} (SEN) measures the proportion of true edges correctly identified (true positives), while \emph{Specificity} (SPE) measures the proportion of absent edges correctly excluded (true negatives).  Finally, we report the \emph{runtime} of each method. All metrics are computed over 50 replications for each $(p, n, s)$ configuration, and summarized by their median values.

\begin{table}
\centering
\caption{Median sensitivity across simulation settings varying sample size, sparsity, and number of variables; bold highlights the best-performing method in each configuration.\label{tab:sens}}
\footnotesize
\renewcommand{\arraystretch}{1.05}
\begin{tabular}{lcccccccc}
\toprule
\textbf{Method} & \multicolumn{4}{c}{\textbf{s = 2}} & \multicolumn{4}{c}{\textbf{s = 5}} \\
\cmidrule(lr){2-5} \cmidrule(lr){6-9}
& 500 & 1000 & 2500 & 5000 & 500 & 1000 & 2500 & 5000 \\
\midrule
\multicolumn{9}{l}{\textbf{p = 10}} \\
copula PC                     & 0.00 & 0.13 & 0.29 & 0.38 & 0.09 & 0.14 & 0.23 & 0.28 \\
latent PC                     & 0.39 & 0.41 & 0.50 & 0.56 & 0.17 & 0.21 & 0.26 & \textbf{0.32} \\
copula score-based            & 0.44 & 0.46 & 0.46 & 0.46 & 0.25 & 0.32 & 0.35 & 0.39 \\
SEM score-based               & 0.50 & 0.58 & 0.67 & 0.75 & \textbf{0.35} & \textbf{0.41} & \textbf{0.48} & \textbf{0.50} \\
latent score-based            & 0.55 & \textbf{0.65} & \textbf{0.67} & 0.72 & \textbf{0.40} & \textbf{0.44} & 0.45 & 0.49 \\
latent MMHC (Gauss, 0.01)      & 0.57 & 0.63 & 0.67 & 0.75 & 0.26 & 0.28 & 0.33 & 0.34 \\
latent MMHC (Gauss, 0.05)      & \textbf{0.60} & 0.63 & \textbf{0.67} & \textbf{0.75} & 0.28 & \textbf{0.29} & \textbf{0.33} & \textbf{0.36} \\
latent MMHC (SEM, 0.01)        & 0.50 & 0.57 & \textbf{0.67} & \textbf{0.75} & 0.25 & 0.30 & 0.30 & 0.36 \\
latent MMHC (SEM, 0.05)        & 0.50 & \textbf{0.58} & \textbf{0.67} & \textbf{0.75} & 0.25 & 0.30 & 0.32 & 0.36 \\
\addlinespace
\midrule
\multicolumn{9}{l}{\textbf{p = 30}} \\
copula PC                     & 0.11 & 0.23 & 0.33 & 0.46 & 0.12 & 0.22 & 0.34 & 0.43 \\
latent PC                     & 0.41 & 0.48 & 0.58 & 0.62 & 0.27 & 0.39 & 0.44 & 0.50 \\
copula score-based            & 0.46 & 0.49 & 0.50 & 0.55 & 0.37 & 0.40 & 0.45 & 0.46 \\
SEM score-based               & 0.54 & 0.67 & 0.74 & 0.78 & \textbf{0.46} & \textbf{0.53} & \textbf{0.61} & \textbf{0.65} \\
latent score-based            & 0.56 & 0.64 & 0.69 & 0.72 & \textbf{0.56} & 0.62 & 0.63 & 0.65 \\
latent MMHC (Gauss, 0.01)      & 0.60 & 0.67 & 0.77 & 0.77 & 0.37 & 0.47 & 0.52 & 0.56 \\
latent MMHC (Gauss, 0.05)      & \textbf{0.60} & \textbf{0.67} & \textbf{0.75} & \textbf{0.75} & 0.40 & \textbf{0.51} & \textbf{0.53} & \textbf{0.56} \\
latent MMHC (SEM, 0.01)        & 0.53 & 0.65 & 0.74 & 0.77 & 0.34 & 0.43 & 0.50 & 0.53 \\
latent MMHC (SEM, 0.05)        & 0.54 & 0.65 & \textbf{0.74} & \textbf{0.77} & 0.36 & 0.44 & 0.49 & 0.54 \\
\bottomrule
\end{tabular}
\end{table}

\subsection{Results}

The structural Hamming distance (SHD) results in Table~\ref{tab:shd} show that our proposed latent MMHC methods consistently outperform existing approaches across all settings. For small graphs ($p = 10$), the SEM-based variants achieve the lowest SHD overall, with the best performance obtained for $\alpha = 0.01$ at small sample sizes and $\alpha = 0.05$ at larger ones. In larger graphs ($p = 30$), the advantage becomes even more pronounced: \emph{latent MMHC (SEM, 0.01)} achieves the lowest SHD in 7 out of 8 settings, with further improvements observed when increasing the PC threshold to $\alpha = 0.05$ in low-sample regimes.

The sensitivity results in Table~\ref{tab:sens} confirm this trend. Both score-based and hybrid methods substantially outperform copula PC and latent PC in terms of edge recovery. The SEM score-based method performs particularly well in dense settings, but is dominated by the hybrid MMHC variants in sparse cases. For both scoring strategies, increasing the PC threshold to $\alpha = 0.05$ yields higher sensitivity, especially when $p = 30$, suggesting that relaxing the constraint-based phase improves edge recall.

\begin{table}
\centering
\caption{Median specificity across simulation settings varying sample size, sparsity, and number of variables; bold highlights the best-performing method in each configuration .\label{tab:spec}}
\footnotesize
\renewcommand{\arraystretch}{1.05}
\begin{tabular}{lcccccccc}
\toprule
\textbf{Method} & \multicolumn{4}{c}{\textbf{s = 2}} & \multicolumn{4}{c}{\textbf{s = 5}} \\
\cmidrule(lr){2-5} \cmidrule(lr){6-9}
& 500 & 1000 & 2500 & 5000 & 500 & 1000 & 2500 & 5000 \\
\midrule
\multicolumn{9}{l}{\textbf{p = 10}} \\
copula PC                     & 0.93 & 0.91 & 0.91 & 0.92 & 0.85 & 0.83 & 0.82 & 0.82 \\
latent PC                     & 0.91 & 0.92 & 0.92 & 0.92 & 0.87 & 0.86 & 0.85 & 0.87 \\
copula score-based            & 0.94 & 0.94 & 0.92 & 0.91 & 0.81 & 0.79 & 0.74 & 0.72 \\
SEM score-based               & 0.95 & 0.94 & 0.94 & 0.94 & 0.83 & 0.80 & 0.76 & 0.74 \\
latent score-based            & 0.86 & 0.88 & 0.90 & 0.90 & 0.67 & 0.65 & 0.65 & 0.65 \\
latent MMHC (Gauss, 0.01)      & 0.95 & 0.95 & 0.95 & 0.95 & 0.91 & 0.88 & 0.89 & 0.88 \\
latent MMHC (Gauss, 0.05)      & 0.93 & 0.94 & 0.94 & 0.94 & 0.90 & 0.87 & 0.88 & 0.86 \\
latent MMHC (SEM, 0.01)        & 0.94 & 0.95 & 0.94 & 0.94 & 0.90 & 0.89 & 0.89 & 0.87 \\
latent MMHC (SEM, 0.05)        & \textbf{0.96} & \textbf{0.96} & \textbf{0.96} & \textbf{0.97} & \textbf{0.91} & \textbf{0.90} & \textbf{0.88} & \textbf{0.88} \\
\addlinespace
\midrule
\multicolumn{9}{l}{\textbf{p = 30}} \\
copula PC                     & 0.97 & 0.97 & 0.97 & 0.97 & 0.96 & 0.96 & 0.95 & 0.95 \\
latent PC                     & 0.97 & 0.97 & 0.97 & 0.97 & 0.96 & 0.97 & 0.97 & 0.97 \\
copula score-based            & 0.97 & 0.97 & 0.97 & 0.97 & 0.93 & 0.92 & 0.91 & 0.89 \\
SEM score-based               & 0.97 & 0.98 & 0.98 & 0.97 & 0.94 & 0.93 & 0.91 & 0.89 \\
latent score-based            & 0.85 & 0.87 & 0.90 & 0.92 & 0.72 & 0.74 & 0.75 & 0.75 \\
latent MMHC (Gauss, 0.01)      & 0.97 & 0.97 & 0.98 & 0.98 & 0.97 & 0.98 & 0.98 & 0.98 \\
latent MMHC (Gauss, 0.05)      & 0.96 & 0.96 & 0.97 & 0.97 & 0.97 & 0.97 & 0.97 & 0.97 \\
latent MMHC (SEM, 0.01)        & \textbf{0.99} & \textbf{0.99} & \textbf{0.99} & \textbf{0.99} & \textbf{0.98} & \textbf{0.98} & \textbf{0.98} & \textbf{0.98} \\
latent MMHC (SEM, 0.05)        & \textbf{0.99} & \textbf{0.99} & \textbf{0.99} & 0.99 & 0.98 & \textbf{0.98} & 0.98 & 0.98 \\
\bottomrule
\end{tabular}
\end{table}

Table~\ref{tab:spec} shows that all MMHC methods maintain high specificity across conditions, often matching or exceeding copula PC and latent PC. In particular, \emph{latent MMHC (SEM, 0.05)} achieves the highest specificity across nearly all settings when $p = 10$, and both SEM-based variants reach the best performance when $p = 30$. These results suggest that increasing the PC threshold does not lead to a significant loss in false positive control.

Finally, Table~\ref{tab:runtime} highlights the computational trade-offs. Constraint-based and score-based methods are generally faster, with copula PC and latent score-based being the most efficient. The hybrid MMHC variants incur higher runtimes, particularly the SEM-based methods when $p = 30$ and $n$ is large. However, their superior structural accuracy and stability make them a competitive option for moderate to large-scale applications.

\begin{table}
\centering
\caption{Median runtime (in seconds) across simulation settings varying sample size, sparsity, and number of variables; bold highlights the best-performing method in each configuration. Runtimes refer only to the DAG learning step, excluding the construction of the latent covariance matrix, in order to provide a fair comparison across methods.\label{tab:runtime}}
\footnotesize
\renewcommand{\arraystretch}{1.05}
\begin{tabular}{lcccccccc}
\toprule
\textbf{Method} & \multicolumn{4}{c}{\textbf{s = 2}} & \multicolumn{4}{c}{\textbf{s = 5}} \\
\cmidrule(lr){2-5} \cmidrule(lr){6-9}
& 500 & 1000 & 2500 & 5000 & 500 & 1000 & 2500 & 5000 \\
\midrule
\multicolumn{9}{l}{\textbf{p = 10}} \\
copula PC                     & \textbf{0.01} & 0.01 & 0.01 & 0.02 & 0.03 & 0.05 & 0.09 & 0.12 \\
latent PC                     & 0.02 & 0.02 & 0.02 & 0.03 & 0.03 & 0.05 & 0.07 & 0.08 \\
copula score-based            & 0.01 & \textbf{0.01} & \textbf{0.01} & \textbf{0.01} & \textbf{0.01} & \textbf{0.01} & \textbf{0.01} & \textbf{0.02} \\
SEM score-based               & 0.11 & 0.13 & 0.22 & 0.38 & 0.14 & 0.19 & 0.33 & 0.62 \\
latent score-based            & 0.01 & 0.01 & \textbf{0.01} & 0.01 & 0.02 & 0.02 & 0.02 & 0.02 \\
latent MMHC (Gauss, 0.01)      & 0.02 & 0.02 & 0.02 & 0.03 & 0.04 & 0.06 & 0.07 & 0.09 \\
latent MMHC (Gauss, 0.05)      & 0.03 & 0.03 & 0.04 & 0.04 & 0.05 & 0.06 & 0.09 & 0.12 \\
latent MMHC (SEM, 0.01)        & 0.05 & 0.06 & 0.10 & 0.17 & 0.08 & 0.11 & 0.17 & 0.28 \\
latent MMHC (SEM, 0.05)        & 0.06 & 0.07 & 0.11 & 0.18 & 0.09 & 0.12 & 0.20 & 0.32 \\
\addlinespace
\midrule
\multicolumn{9}{l}{\textbf{p = 30}} \\
copula PC                     & \textbf{0.05} & 0.07 & 0.12 & 0.20 & 0.26 & 0.61 & 1.73 & 3.71 \\
latent PC                     & 0.42 & 0.75 & 0.58 & 0.59 & 0.37 & 0.61 & 0.98 & 1.20 \\
copula score-based            & 0.06 & \textbf{0.06} & \textbf{0.07} & \textbf{0.07} & \textbf{0.11} & \textbf{0.12} & \textbf{0.12} & \textbf{0.14} \\
SEM score-based               & 0.81 & 1.01 & 1.64 & 2.69 & 1.22 & 1.70 & 3.31 & 6.72 \\
latent score-based            & 0.18 & 0.16 & 0.14 & 0.12 & 0.31 & 0.31 & 0.29 & 0.29 \\
latent MMHC (Gauss, 0.01)      & 0.44 & 0.76 & 0.64 & 0.61 & 0.39 & 0.66 & 0.95 & 1.25 \\
latent MMHC (Gauss, 0.05)      & 1.02 & 2.16 & 2.19 & 2.11 & 0.63 & 0.97 & 1.30 & 1.64 \\
latent MMHC (SEM, 0.01)        & 0.55 & 0.88 & 0.88 & 1.09 & 0.50 & 0.79 & 1.32 & 1.87 \\
latent MMHC (SEM, 0.05)        & 1.04 & 2.27 & 2.37 & 2.53 & 0.74 & 1.13 & 1.59 & 2.29 \\
\bottomrule
\end{tabular}
\end{table}

Overall, the results show that latent MMHC methods, particularly those based on the SEM score, offer a robust and scalable approach to structure learning in mixed data. They outperform copula PC and latent PC in both sensitivity and structural accuracy, while maintaining strong specificity and acceptable runtime.

\begin{figure}
    \centering
    \includegraphics[width=0.8\linewidth]{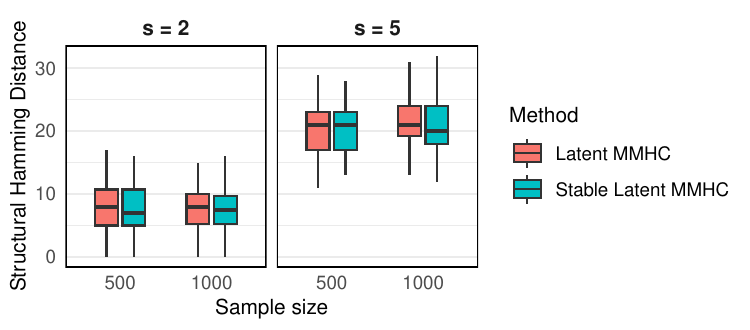}
    \caption{SHD for latent MMHC and its stable (bootstrap-aggregated) variant across sample sizes and sparsity levels, for $p = 10$. Each panel corresponds to a different sparsity setting ($s = 2$ on the left and $s = 5$ on the right), and boxplots compare the standard and stable versions at $n = 500$ and $n = 1000$.}
    \label{fig:p1_shd_comparison}
\end{figure}

\subsection{Assessing the Stable Latent MMHC Variant}
We conducted a focused simulation to evaluate the potential benefits of bootstrap aggregation in our setting. As the bootstrap procedure is computationally intensive, we restricted the analysis to the case $p=10$, with sample sizes $n\in\{500,1000\}$. For the stable variant, we applied $B=200$ bootstrap resamples within each replicate.
We focused on the latent MMHC SEM model with $\alpha=0.01$, as it showed the best overall performance in the full simulation study. We compared this baseline against its bootstrap-aggregated (stable) counterpart.

Figure~\ref{fig:p1_shd_comparison} summarizes the SHD results. The bootstrap-aggregated stable latent MMHC shows only modest gains over the baseline: improvements are visible either in slightly lower medians or in reduced variability across replications, depending on the configuration. The effect is not dramatic, but the stable variant generally delivers more consistent performance, in line with the idea that aggregation can dampen spurious edges and reinforce those that recur across resamples
.
\section{Modeling Psychological Traits of Volleyball Athletes}
\label{sec:volley}

\subsection{Data}

The aim of our application is to explore how psychological traits and background factors combine to shape the profiles of competitive volleyball players. To this end, we draw on a dataset of 164 female athletes competing in the C and D leagues of the Campania region in Italy collected within the Laboratory for Statistical Data Analysis (LASt) – University of Naples Federico II, Italy, in 2020. Alongside demographic and sport-related information (e.g., age group, role on court, weekly training volume, competition level), the study collected three well-established psychological measures.

Personality traits were assessed through the NEO Five-Factor Inventory \citep{mccrae2008fivefactor}, a 15-item short form that evaluates the Big Five dimensions: extraversion, agreeableness, conscientiousness, neuroticism (emotional stability), and openness. Mental skills were measured using the sport performance psychological inventory (IPPS-48) \citep{robazza2009ipps}, which covers eight psychological skill areas: self-talk, goal setting, self-confidence, emotional arousal control, cognitive anxiety, concentration disruption, mental practice, and race preparation. Finally, self-esteem was evaluated with the Rosenberg Self-Esteem Scale \citep{rosenberg1965selfimage}, a 10-item uni-dimensional scale capturing global self-worth. 

For each construct, we computed the average of the underlying item responses, obtaining continuous scores that represent the latent trait rather than individual ordinal answers. This aggregation is standard practice in psychometrics and reflects the assumption that the constructs are inherently continuous. Together with demographic and experience variables, these measures form a heterogeneous dataset with continuous, ordinal and binary information. Table~\ref{tab:volley_bn_vars} summarizes the variables considered in the analysis.

\begin{table*}
\caption{List of variables used in the BN analysis of volleyball athletes, with descriptions and states. Continuous variables correspond to averaged psychometric scores and retain their original ordinal range.}
\label{tab:volley_bn_vars}
\footnotesize
\centering
\scalebox{0.78}{
\renewcommand{\arraystretch}{1.15}
\begin{tabular}{p{2.1cm}p{2.5cm}p{5cm}p{5cm}}
\toprule
\textbf{Group} & \textbf{Variable} & \textbf{Description} & \textbf{States / Range} \\
\midrule
\multirow{6}{*}{\textbf{Sports}} 
& league\_division & Competition level of player's team & C, D \\
& num\_competitions & Number of official competitions played & None, 1–3, 4–10, 11–20, $>$20 \\
& player\_position & Volleyball playing position & Non\_Hitter, Hitter \\
& plays\_other\_sports & Whether the player practices other sports & No, Yes \\
& referee\_experience & Whether the player has experience as a referee & No, Yes \\
& weekly\_training\_hours & Weekly training volume & $<$3h, 3–6h, 6–9h, $>$9h \\
\midrule
\multirow{5}{*}{\textbf{Lifestyle}} 
& age\_group & Age group of respondent & 18–19, 20–24, 25–29, 30+ \\
& drinks\_alcohol & Alcohol consumption status & No, Yes \\
& in\_relationship & Whether the player is in a relationship & No, Yes \\
& working\_status & Main life activity & Not\_Working, Working \\
& smokes & Smoking status & No, Yes \\
\midrule
\multirow{5}{*}{\textbf{Big Five}} 
& agreeableness & Compassion, helpfulness, and cooperation & Continuous average, original range 1–5 \\
& conscientiousness & Self-control, diligence, attention to detail & Continuous average, original range 1–5 \\
& extraversion & Boldness, energy, and social interactivity & Continuous average, original range 1–5 \\
& neuroticism & Moodiness, irritability, and emotional instability & Continuous average, original range 1–5 \\
& openness & Curiosity, creativity, and openness to new ideas & Continuous average, original range 1–5 \\
\midrule
\textbf{Self-Esteem} 
& self\_esteem & Global evaluation of self-worth & Continuous average, original range 1–4 \\
\midrule
\multirow{8}{*}{\textbf{IPPS-48}} 
& concentration\_issues & Difficulties in maintaining attentional focus & Continuous average, original range 1–6 \\
& emotional\_arousal & Regulation of emotional activation during play & Continuous average, original range 1–6 \\
& goal\_setting & Tendency to define and pursue performance goals & Continuous average, original range 1–6 \\
& match\_preparation & Strategies used before competition for mental readiness & Continuous average, original range 1–6 \\
& mental\_practice & Use of imagery or mental rehearsal strategies & Continuous average, original range 1–6 \\
& self\_confidence & Belief in own performance capacity & Continuous average, original range 1–6 \\
& self\_talk & Use of internal or external speech for focus & Continuous average, original range 1–6 \\
& worry & Cognitive and emotional concern during performance & Continuous average, original range 1–6 \\
\bottomrule
\end{tabular}}
\end{table*}

\subsection{Model Learning}

To estimate the network structure, we applied the stable latent MMHC algorithm with the SEM score and significance level $\alpha = 0.01$. Given the relatively small sample size, we increased the number of bootstrap replicates to $B=500$. In addition, edges from psychological traits to lifestyle and sport-related background variables were forbidden, as the latter are better understood as exogenous covariates.

The consensus structure obtained from the bootstrap aggregation step was then used as the backbone of the final network. At this stage, the continuous trait scores were discretized back to their original ordinal scales (e.g., 1--5 or 1--6) by rounding to the nearest category. A fully discrete BN was then fitted using Bayesian parameter estimation with a Dirichlet prior equal to one for all conditional probability tables. The Dirichlet prior acts as a regularization device, which is especially important given the limited sample size: it avoids zero probabilities in the estimated tables, improves the robustness of inferences, and supports more trustworthy out-of-sample predictions. This is a standard choice in discrete BN learning, ensuring stable estimation across different applications. This two-step strategy leverages the information in continuous scores during structure learning while ensuring interpretability of the final model in terms of the original psychological categories and enabling the use of standard discrete BN inference tools.

\subsection{Results}

\begin{figure}[t]
    \centering
    \includegraphics[width=0.95\linewidth]{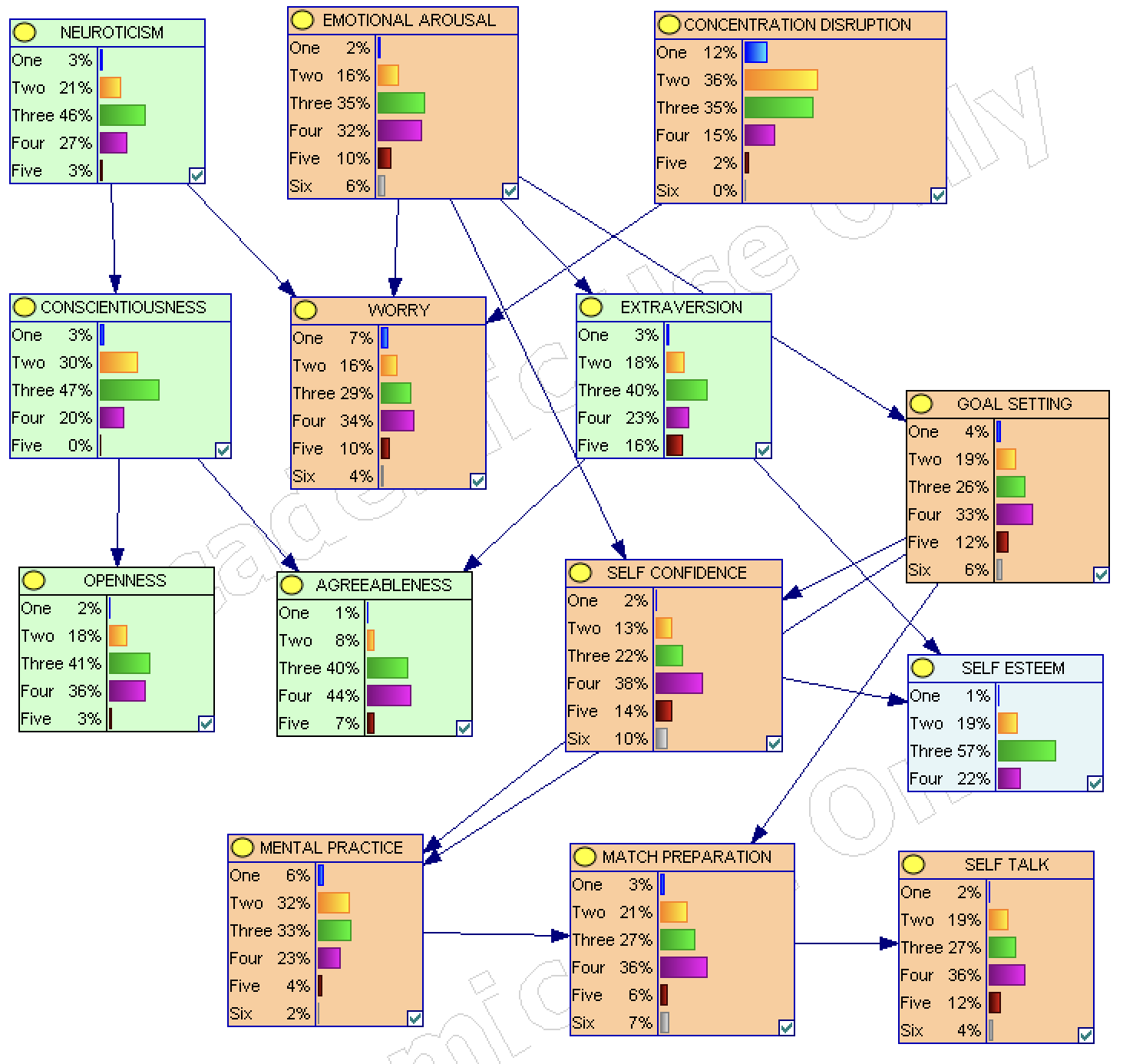}
    \caption{Consensus BN of psychological traits for volleyball athletes, visualized using GeNIe. 
    Nodes are color-coded by construct: green for Big Five traits, orange for IPPS-48 skills, 
    and light blue for self-esteem. Each node includes a bar plot reporting its marginal 
    distribution as computed from the estimated model.}
    \label{fig:volley_net}
\end{figure}

The final network is shown in Figure~\ref{fig:volley_net} and is freely available within the \texttt{bnRep} R package \citep{leonelli2025bnrep}. None of the sport- or lifestyle-related variables were connected to the psychological traits, and for this reason they are omitted. The structure therefore focuses exclusively on the interrelations among the psychological constructs.

Several patterns stand out. Personality traits appear as upstream factors: \textit{neuroticism} influences both \textit{conscientiousness} and \textit{worry}, while \textit{conscientiousness} further drives \textit{openness} and \textit{agreeableness}. \textit{Extraversion} is linked to \textit{agreeableness} and to higher \textit{self-esteem}. Together these pathways suggest that the Big Five, and in particular neuroticism, play a central role in shaping both other traits and anxiety-related outcomes.

Within the mental skills cluster, \textit{goal setting} and \textit{self-confidence} occupy a central position. \textit{Goal setting} predicts \textit{self-confidence}, and propagates to \textit{mental practice} and \textit{match preparation}, the latter of which leads to greater use of \textit{self-talk}. \textit{Self-confidence} also feeds directly into \textit{self-esteem}, highlighting its role as a pivotal node bridging performance-related skills and global self-evaluations.

Finally, \textit{emotional arousal} emerges as an important connector, connected to \textit{extraversion}, \textit{goal setting}, \textit{self-confidence}, and \textit{worry}. Together with the influence of \textit{concentration disruption} on \textit{worry}, this underscores the role of emotion regulation and attentional control in shaping anxiety responses.

Overall, the learned network points to a layered structure in which basic personality traits shape coping-related constructs such as \textit{worry}, while motivational and preparatory skills form an interconnected subsystem organized around \textit{goal setting} and \textit{self-confidence}.

\subsection{Sensitivity Analysis}

To quantify the relative importance of predictors for key downstream variables, 
we computed variance-based Sobol indices using the exact algorithm introduced in \citep{ballester2022computing}. 
Indices represent the proportion of variance in the output explained by each input. 
Table~\ref{tab:sobol} reports the results for four outputs of particular interest: 
\emph{Worry}, \emph{Self-confidence}, \emph{Match preparation}, and \emph{Self-esteem}. 

\begin{table}
\caption{Sobol indices (in \%) for four selected outputs. 
Entries correspond to the variance component attributed to each input variable; 
zeros indicate negligible contributions.}
\centering
\footnotesize
\renewcommand{\arraystretch}{1.1}
\begin{tabular}{lcccc}
\toprule
\textbf{Input} & \textbf{Worry} & \textbf{Self-confidence} & \textbf{Match preparation} & \textbf{Self-esteem} \\
\midrule
Neuroticism              & 6.95 & 0.00 & 0.00 & 0.00 \\
Agreeableness            & 0.16 & 0.65 & 0.19 & 2.58 \\
Conscientiousness        & 0.38 & 0.00 & 0.00 & 0.00 \\
Extraversion             & 2.85 & 12.88 & 3.86 & 61.41 \\
Openness                 & 0.05 & 0.00 & 0.00 & 0.00 \\
Self-talk                & 0.23 & 5.64 & 55.59 & 1.92 \\
Goal setting             & 3.81 & 61.86 & 38.26 & 22.97 \\
Self-confidence          & 3.73 & --   & 31.27 & 45.56 \\
Emotional arousal        & 14.55 & 49.05 & 14.72 & 38.49 \\
Worry                    & --   & 1.85 & 0.56 & 1.41 \\
Concentration disruption & 24.47 & 0.00 & 0.00 & 0.00 \\
Mental practice          & 1.34 & 37.35 & 35.75 & 12.99 \\
Match preparation        & 0.83 & 20.28 & --   & 7.01 \\
Self-esteem              & 1.04 & 25.14 & 5.60 & --   \\
\bottomrule
\end{tabular}
\label{tab:sobol}
\end{table}

The decomposition highlights distinct drivers for each construct. 
\emph{Worry} is dominated by concentration disruption, with further contributions from emotional arousal and neuroticism. 
\emph{Self-confidence} is primarily shaped by goal setting and emotional arousal, with additional influence from mental practice. 
\emph{Match preparation} receives balanced contributions from goal setting and mental practice, while self-talk appears almost entirely explained by preparation. 
Finally, \emph{Self-esteem} reflects both personality (extraversion) and skills (self-confidence, emotional arousal), suggesting that athletes’ global self-worth arises from an interaction of dispositional and trainable factors. 
Overall, these results confirm the layered structure identified in the network and quantify the extent to which each factor accounts for variability in key outcomes.

\subsection{Conditional Probabilities and Belief Dynamics}

While Sobol indices identify the most influential predictors of a target, they do not reveal 
the shape or direction of these effects. In a BN framework, this information is 
captured by conditional probabilities, which describe how beliefs about an outcome update 
when evidence on a precursor is observed. To illustrate, we focus on \emph{Match preparation}, 
one of the key downstream skills identified in the network. The corresponding conditional 
probabilities are reported in Table~\ref{tab:mp_shift}, while results for other outputs are 
deferred to the Appendix.

The table shows the baseline distribution of \emph{Match preparation} alongside the conditional 
distributions when each of the five Sobol-relevant predictors (\emph{Self-talk}, \emph{Goal setting}, 
\emph{Self-confidence}, \emph{Emotional arousal}, and \emph{Mental practice}) is fixed to one of its 
ordinal levels. Clear monotonic patterns emerge. For \emph{Self-talk} and \emph{Self-confidence}, very 
low levels (One–Two) concentrate mass on poor preparation, while higher levels (Four–Six) 
shift probability toward strong preparation. \emph{Goal setting} and \emph{Mental practice} display 
a similar gradient, with their higher levels producing the sharpest increases in the top 
preparation states. \emph{Emotional arousal} has a more gradual effect, but higher levels still 
reduce the probability of poor preparation and increase the likelihood of mid-to-high preparation.

Taken together, these conditional distributions complement the variance-based analysis by 
clarifying not only which skills matter most, but also how their improvement reshapes the 
distribution of preparation outcomes. More generally, the additional results reported in the 
Appendix confirm that this pattern holds across other psychological targets, underscoring 
the value of conditional probability analysis as a complement to variance decomposition.

\begin{table*}[t]
\centering
\footnotesize
\caption{Conditional distributions of \emph{Match preparation} (columns; categories One–Six) given key precursors (rows), compared to the baseline distribution.}
\label{tab:mp_shift}
\setlength{\tabcolsep}{5.5pt}
\begin{tabular}{lrrrrrr}
\toprule
 & One & Two & Three & Four & Five & Six \\
\midrule
\multicolumn{7}{l}{\textbf{Baseline (no evidence)}}\\
Baseline & 0.03 & 0.21 & 0.27 & 0.36 & 0.06 & 0.07 \\
\addlinespace[4pt]

\multicolumn{7}{l}{\textbf{Given Self-talk}}\\
One   & 0.32 & 0.64 & 0.01 & 0.01 & 0.01 & 0.01 \\
Two   & 0.06 & 0.47 & 0.37 & 0.09 & 0.00 & 0.00 \\
Three & 0.02 & 0.20 & 0.29 & 0.41 & 0.07 & 0.00 \\
Four  & 0.02 & 0.12 & 0.25 & 0.47 & 0.07 & 0.07 \\
Five  & 0.00 & 0.05 & 0.21 & 0.42 & 0.11 & 0.21 \\
Six   & 0.00 & 0.00 & 0.00 & 0.42 & 0.14 & 0.42 \\
\addlinespace[4pt]

\multicolumn{7}{l}{\textbf{Given Goal setting}}\\
One   & 0.17 & 0.33 & 0.33 & 0.17 & 0.00 & 0.00 \\
Two   & 0.10 & 0.32 & 0.32 & 0.23 & 0.00 & 0.03 \\
Three & 0.02 & 0.32 & 0.30 & 0.23 & 0.09 & 0.02 \\
Four  & 0.00 & 0.13 & 0.24 & 0.56 & 0.04 & 0.04 \\
Five  & 0.00 & 0.00 & 0.26 & 0.37 & 0.16 & 0.21 \\
Six   & 0.00 & 0.10 & 0.10 & 0.40 & 0.10 & 0.30 \\
\addlinespace[4pt]

\multicolumn{7}{l}{\textbf{Given Self-confidence}}\\
One   & 0.20 & 0.36 & 0.29 & 0.13 & 0.01 & 0.01 \\
Two   & 0.09 & 0.29 & 0.30 & 0.25 & 0.03 & 0.03 \\
Three & 0.04 & 0.33 & 0.27 & 0.28 & 0.04 & 0.03 \\
Four  & 0.01 & 0.14 & 0.28 & 0.49 & 0.06 & 0.02 \\
Five  & 0.01 & 0.19 & 0.31 & 0.33 & 0.08 & 0.09 \\
Six   & 0.00 & 0.07 & 0.10 & 0.32 & 0.17 & 0.33 \\
\addlinespace[4pt]

\multicolumn{7}{l}{\textbf{Given Emotional arousal}}\\
One   & 0.11 & 0.38 & 0.31 & 0.15 & 0.03 & 0.02 \\
Two   & 0.07 & 0.31 & 0.28 & 0.24 & 0.04 & 0.04 \\
Three & 0.04 & 0.23 & 0.29 & 0.35 & 0.05 & 0.04 \\
Four  & 0.01 & 0.16 & 0.27 & 0.44 & 0.06 & 0.06 \\
Five  & 0.01 & 0.13 & 0.22 & 0.39 & 0.10 & 0.16 \\
Six   & 0.00 & 0.08 & 0.19 & 0.38 & 0.13 & 0.22 \\
\addlinespace[4pt]

\multicolumn{7}{l}{\textbf{Given Mental practice}}\\
One   & 0.11 & 0.22 & 0.55 & 0.11 & 0.00 & 0.00 \\
Two   & 0.06 & 0.44 & 0.23 & 0.17 & 0.02 & 0.08 \\
Three & 0.02 & 0.13 & 0.31 & 0.44 & 0.09 & 0.02 \\
Four  & 0.00 & 0.03 & 0.26 & 0.60 & 0.03 & 0.08 \\
Five  & 0.00 & 0.14 & 0.00 & 0.28 & 0.28 & 0.28 \\
Six   & 0.01 & 0.01 & 0.01 & 0.32 & 0.32 & 0.32 \\
\bottomrule
\end{tabular}
\end{table*}

\subsection{Joint Scenario Reasoning}

While one-at-a-time conditioning clarifies the marginal impact of individual traits, 
practitioners are often interested in the combined effect of multiple skills or dispositions. 
BN allow for such \emph{joint scenario reasoning}, where evidence is set 
on several predictors simultaneously and the resulting outcome distributions are compared 
to the baseline. This approach provides a decision-oriented complement to variance-based 
sensitivity analysis.

Table~\ref{tab:scenarios} summarizes six illustrative profiles of volleyball athletes 
considered in our analysis. These scenarios were constructed by fixing combinations of 
psychological traits suggested by the network structure and sensitivity analysis, and 
represent distinct motivational and cognitive styles.

\begin{table}[h!]
\centering
\footnotesize
\caption{Definition of illustrative scenarios used for joint reasoning.}
\label{tab:scenarios}
\begin{tabular}{ll}
\toprule
\textbf{Scenario} & \textbf{Definition (fixed evidence)} \\
\midrule
High motivation \& arousal & GOAL SETTING = Six, EMOTIONAL AROUSAL = Six \\
Low motivation \& arousal  & GOAL SETTING = One, EMOTIONAL AROUSAL = One \\
Skills boost               & SELF TALK = Five, MENTAL PRACTICE = Five \\
Rehearsal focus            & MENTAL PRACTICE = Six, GOAL SETTING = Four \\
Resilient extrovert        & EXTRAVERSION = Five, NEUROTICISM = Two \\
Vulnerable attentional     & NEUROTICISM = Five, CONCENTRATION DISRUPTION = Five \\
\bottomrule
\end{tabular}
\end{table}

\begin{figure}[t]
    \centering
    \includegraphics[width=\linewidth]{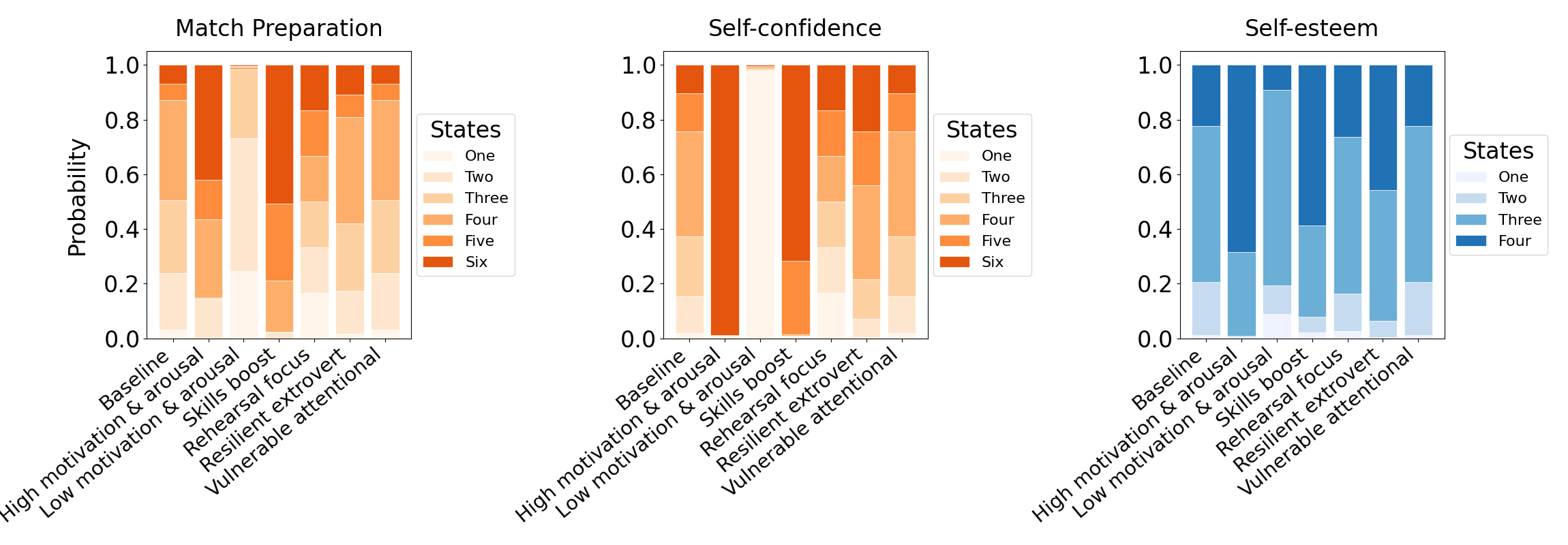}
    \caption{Scenario-conditional distributions of \emph{Match preparation}, 
    \emph{Self-confidence}, and \emph{Self-esteem}. Bars show the baseline and 
    six illustrative athlete profiles defined in Table~\ref{tab:scenarios}.}
    \label{fig:scenario_barplots}
\end{figure}

Figure~\ref{fig:scenario_barplots} displays the scenario-conditional distributions 
for three downstream outcomes of particular interest: \emph{Match preparation}, 
\emph{Self-confidence}, and \emph{Self-esteem}. Each bar summarizes the probability 
distribution over ordinal categories, with the baseline shown alongside the six scenarios.

Clear contrasts emerge across profiles. Athletes in the \emph{High motivation \& arousal} 
scenario are strongly shifted toward the top categories of both self-confidence and 
self-esteem, with match preparation also redistributed toward higher readiness. 
Conversely, the \emph{Low motivation \& arousal} scenario concentrates mass in the lowest 
states of self-confidence and preparation, underscoring the joint importance of goal 
orientation and arousal regulation. The \emph{Skills boost} and \emph{Rehearsal focus} 
profiles also increase the likelihood of stronger preparation, albeit through different 
combinations of mental skills. Finally, the \emph{Resilient extrovert} profile improves 
both preparation and confidence, while the \emph{Vulnerable attentional} scenario closely 
resembles the baseline, reflecting limited incremental impact when neuroticism and 
concentration disruption are both high.

\section{Conclusions}
\label{sec:conc}

This paper introduced \emph{latent MMHC}, a new hybrid algorithm for learning BNs with mixed-type variables. By combining a latent Gaussian copula representation with a constraint-based skeleton and a constrained score-based refinement, the method extends existing copula-based learners to return a single oriented DAG. A bootstrap-aggregated variant further enhances stability. Together, these advances provide a general framework for extracting interpretable graphical structures from heterogeneous data, bridging a gap between purely discrete and Gaussian approaches.

Through a comprehensive simulation study, we showed that latent MMHC outperforms established competitors across a range of conditions. In particular, the SEM-based scoring variants reduced structural Hamming distance and improved edge recall, while maintaining high specificity and acceptable runtimes. These results confirm that a hybrid approach offers tangible benefits over constraint-only procedures, especially in orienting edges reliably under finite samples.

We then applied the method to a dataset of 164 female volleyball athletes, combining demographic information with standardized psychological measures. Treating questionnaire responses as continuous in the learning phase and then discretizing back to ordinal scales for interpretation proved effective: the procedure captured nuanced dependencies during structure estimation while yielding a final model that was interpretable in terms of the original psychometric categories. This approach illustrates a principled workflow for sports psychology datasets, where constructs are inherently latent continuous traits but are commonly measured on ordinal scales.

The learned network of psychological attributes revealed a layered organization: Big Five traits, and in particular neuroticism and extraversion, appeared upstream of coping-related constructs; goal setting and self-confidence formed a tightly linked motivational core; and emotional arousal connected motivation to anxiety-related outcomes. Sobol indices and conditional probability analyses quantified how shifts in key skills, such as improving goal setting or mental practice, propagate through the system to enhance match preparation, self-confidence, and ultimately self-esteem.

From an applied perspective, these findings underscore the potential of BNs as practical tools for coaches, sport psychologists, and athlete support staff. By modeling conditional relations explicitly, BNs allow practitioners to move beyond single-skill assessments and to reason about how improvements in one area may ripple through others. Scenario analyses further provide an intuitive interface for exploring “what-if” questions, making BNs a promising complement to traditional profiling methods.

More broadly, the study illustrates how routinely collected psychological data can be transformed into decision support tools. Coaches can use such models to identify leverage points in mental skills training, to tailor interventions to individual profiles, and to anticipate indirect effects that may otherwise be overlooked. Incorporating probabilistic reasoning into athlete development thus has the potential to enhance both the personalization and effectiveness of training programs.

Future work may extend this line of research by integrating longitudinal data, linking psychological profiles with performance outcomes, and exploring causal inference under additional assumptions. Nonetheless, the present results demonstrate that BNs, learned via hybrid copula-based methods, offer a transparent and actionable framework for modeling the complex interplay of psychological traits in competitive sport.

\bibliography{sn-bibliography}


\begin{thebibliography}{62}
\ifx \bisbn   \undefined \def \bisbn  #1{ISBN #1}\fi
\ifx \binits  \undefined \def \binits#1{#1}\fi
\ifx \bauthor  \undefined \def \bauthor#1{#1}\fi
\ifx \batitle  \undefined \def \batitle#1{#1}\fi
\ifx \bjtitle  \undefined \def \bjtitle#1{#1}\fi
\ifx \bvolume  \undefined \def \bvolume#1{\textbf{#1}}\fi
\ifx \byear  \undefined \def \byear#1{#1}\fi
\ifx \bissue  \undefined \def \bissue#1{#1}\fi
\ifx \bfpage  \undefined \def \bfpage#1{#1}\fi
\ifx \blpage  \undefined \def \blpage #1{#1}\fi
\ifx \burl  \undefined \def \burl#1{\textsf{#1}}\fi
\ifx \doiurl  \undefined \def \doiurl#1{\url{https://doi.org/#1}}\fi
\ifx \betal  \undefined \def \betal{\textit{et al.}}\fi
\ifx \binstitute  \undefined \def \binstitute#1{#1}\fi
\ifx \binstitutionaled  \undefined \def \binstitutionaled#1{#1}\fi
\ifx \bctitle  \undefined \def \bctitle#1{#1}\fi
\ifx \beditor  \undefined \def \beditor#1{#1}\fi
\ifx \bpublisher  \undefined \def \bpublisher#1{#1}\fi
\ifx \bbtitle  \undefined \def \bbtitle#1{#1}\fi
\ifx \bedition  \undefined \def \bedition#1{#1}\fi
\ifx \bseriesno  \undefined \def \bseriesno#1{#1}\fi
\ifx \blocation  \undefined \def \blocation#1{#1}\fi
\ifx \bsertitle  \undefined \def \bsertitle#1{#1}\fi
\ifx \bsnm \undefined \def \bsnm#1{#1}\fi
\ifx \bsuffix \undefined \def \bsuffix#1{#1}\fi
\ifx \bparticle \undefined \def \bparticle#1{#1}\fi
\ifx \barticle \undefined \def \barticle#1{#1}\fi
\bibcommenthead
\ifx \bconfdate \undefined \def \bconfdate #1{#1}\fi
\ifx \botherref \undefined \def \botherref #1{#1}\fi
\ifx \url \undefined \def \url#1{\textsf{#1}}\fi
\ifx \bchapter \undefined \def \bchapter#1{#1}\fi
\ifx \bbook \undefined \def \bbook#1{#1}\fi
\ifx \bcomment \undefined \def \bcomment#1{#1}\fi
\ifx \oauthor \undefined \def \oauthor#1{#1}\fi
\ifx \citeauthoryear \undefined \def \citeauthoryear#1{#1}\fi
\ifx \endbibitem  \undefined \def \endbibitem {}\fi
\ifx \bconflocation  \undefined \def \bconflocation#1{#1}\fi
\ifx \arxivurl  \undefined \def \arxivurl#1{\textsf{#1}}\fi
\csname PreBibitemsHook\endcsname

\bibitem[\protect\citeauthoryear{Ayranci and Aydin}{2025}]{ayranci2025complex}
\begin{barticle}
\bauthor{\bsnm{Ayranci}, \binits{M.}},
\bauthor{\bsnm{Aydin}, \binits{M.K.}}:
\batitle{The complex interplay between psychological factors and sports performance: A systematic review and meta-analysis}.
\bjtitle{PloS One}
\bvolume{20}(\bissue{8}),
\bfpage{0330862}
(\byear{2025})
\end{barticle}
\endbibitem

\bibitem[\protect\citeauthoryear{Brown and Fletcher}{2017}]{brown2017effects}
\begin{barticle}
\bauthor{\bsnm{Brown}, \binits{D.J.}},
\bauthor{\bsnm{Fletcher}, \binits{D.}}:
\batitle{Effects of psychological and psychosocial interventions on sport performance: A meta-analysis}.
\bjtitle{Sports Medicine}
\bvolume{47}(\bissue{1}),
\bfpage{77}--\blpage{99}
(\byear{2017})
\end{barticle}
\endbibitem

\bibitem[\protect\citeauthoryear{Lochbaum et~al.}{2021}]{lochbaum2021profile}
\begin{botherref}
\oauthor{\bsnm{Lochbaum}, \binits{M.}},
\oauthor{\bsnm{Zanatta}, \binits{T.}},
\oauthor{\bsnm{Kirschling}, \binits{D.}},
\oauthor{\bsnm{May}, \binits{E.}}:
The profile of moods states and athletic performance: A meta-analysis of published studies.
European Journal of Investigation in Health, Psychology and Education
\textbf{11}(1)
(2021)
\end{botherref}
\endbibitem

\bibitem[\protect\citeauthoryear{Allen et~al.}{2013}]{allen2013personality}
\begin{barticle}
\bauthor{\bsnm{Allen}, \binits{M.S.}},
\bauthor{\bsnm{Greenlees}, \binits{I.}},
\bauthor{\bsnm{Jones}, \binits{M.}}:
\batitle{Personality in sport: A comprehensive review}.
\bjtitle{International Review of Sport and Exercise Psychology}
\bvolume{6}(\bissue{1}),
\bfpage{184}--\blpage{208}
(\byear{2013})
\end{barticle}
\endbibitem

\bibitem[\protect\citeauthoryear{Allen and Laborde}{2014}]{allen2014role}
\begin{barticle}
\bauthor{\bsnm{Allen}, \binits{M.S.}},
\bauthor{\bsnm{Laborde}, \binits{S.}}:
\batitle{The role of personality in sport and physical activity}.
\bjtitle{Current Directions in Psychological Science}
\bvolume{23}(\bissue{6}),
\bfpage{460}--\blpage{465}
(\byear{2014})
\end{barticle}
\endbibitem

\bibitem[\protect\citeauthoryear{Laborde et~al.}{2020}]{laborde2020trait}
\begin{barticle}
\bauthor{\bsnm{Laborde}, \binits{S.}},
\bauthor{\bsnm{Allen}, \binits{M.S.}},
\bauthor{\bsnm{Katschak}, \binits{K.}},
\bauthor{\bsnm{Mattonet}, \binits{K.}},
\bauthor{\bsnm{Lachner}, \binits{N.}}:
\batitle{Trait personality in sport and exercise psychology: A mapping review and research agenda}.
\bjtitle{International Journal of Sport and Exercise Psychology}
\bvolume{18}(\bissue{6}),
\bfpage{701}--\blpage{716}
(\byear{2020})
\end{barticle}
\endbibitem

\bibitem[\protect\citeauthoryear{Albert et~al.}{2017}]{albert2017handbook}
\begin{bbook}
\bauthor{\bsnm{Albert}, \binits{J.}},
\bauthor{\bsnm{Glickman}, \binits{M.E.}},
\bauthor{\bsnm{Swartz}, \binits{T.B.}},
\bauthor{\bsnm{Koning}, \binits{R.H.}}:
\bbtitle{Handbook of Statistical Methods and Analyses in Sports}.
\bpublisher{CRC Press},
\blocation{London}
(\byear{2017})
\end{bbook}
\endbibitem

\bibitem[\protect\citeauthoryear{McGarry}{2009}]{mcgarry2009applied}
\begin{barticle}
\bauthor{\bsnm{McGarry}, \binits{T.}}:
\batitle{Applied and theoretical perspectives of performance analysis in sport: Scientific issues and challenges}.
\bjtitle{International Journal of Performance Analysis in Sport}
\bvolume{9}(\bissue{1}),
\bfpage{128}--\blpage{140}
(\byear{2009})
\end{barticle}
\endbibitem

\bibitem[\protect\citeauthoryear{Fabbricatore et~al.}{2023}]{fabbricatore2023component}
\begin{barticle}
\bauthor{\bsnm{Fabbricatore}, \binits{R.}},
\bauthor{\bsnm{Iannario}, \binits{M.}},
\bauthor{\bsnm{Romano}, \binits{R.}},
\bauthor{\bsnm{Vistocco}, \binits{D.}}:
\batitle{Component-based structural equation modeling for the assessment of psycho-social aspects and performance of athletes: Measurement and evaluation of swimmers}.
\bjtitle{AStA Advances in Statistical Analysis}
\bvolume{107}(\bissue{1}),
\bfpage{343}--\blpage{367}
(\byear{2023})
\end{barticle}
\endbibitem

\bibitem[\protect\citeauthoryear{Kalinowski et~al.}{2020}]{kalinowski2020mediational}
\begin{barticle}
\bauthor{\bsnm{Kalinowski}, \binits{P.}},
\bauthor{\bsnm{Bojkowski}, \binits{{\L}.}},
\bauthor{\bsnm{{\'S}liwowski}, \binits{R.}},
\bauthor{\bsnm{Wieczorek}, \binits{A.}},
\bauthor{\bsnm{Konarski}, \binits{J.}},
\bauthor{\bsnm{Tomczak}, \binits{M.}}:
\batitle{Mediational role of coping with stress in relationship between personality and effectiveness of performance of soccer players}.
\bjtitle{International Journal of Sports Science \& Coaching}
\bvolume{15}(\bissue{3}),
\bfpage{354}--\blpage{363}
(\byear{2020})
\end{barticle}
\endbibitem

\bibitem[\protect\citeauthoryear{Iannario et~al.}{2023}]{iannario2023dyadic}
\begin{barticle}
\bauthor{\bsnm{Iannario}, \binits{M.}},
\bauthor{\bsnm{Romano}, \binits{R.}},
\bauthor{\bsnm{Vistocco}, \binits{D.}}:
\batitle{Dyadic analysis for multi-block data in sport surveys analytics}.
\bjtitle{Annals of Operations Research}
\bvolume{325}(\bissue{1}),
\bfpage{701}--\blpage{714}
(\byear{2023})
\end{barticle}
\endbibitem

\bibitem[\protect\citeauthoryear{Koller and Friedman}{2009}]{koller2009probabilistic}
\begin{bbook}
\bauthor{\bsnm{Koller}, \binits{D.}},
\bauthor{\bsnm{Friedman}, \binits{N.}}:
\bbtitle{Probabilistic Graphical Models: Principles and Techniques}.
\bpublisher{MIT Press},
\blocation{London}
(\byear{2009})
\end{bbook}
\endbibitem

\bibitem[\protect\citeauthoryear{Lauritzen}{1996}]{lauritzen1996graphical}
\begin{bbook}
\bauthor{\bsnm{Lauritzen}, \binits{S.L.}}:
\bbtitle{Graphical Models}.
\bpublisher{Clarendon Press},
\blocation{Oxford}
(\byear{1996})
\end{bbook}
\endbibitem

\bibitem[\protect\citeauthoryear{Briganti et~al.}{2023}]{briganti2023tutorial}
\begin{barticle}
\bauthor{\bsnm{Briganti}, \binits{G.}},
\bauthor{\bsnm{Scutari}, \binits{M.}},
\bauthor{\bsnm{McNally}, \binits{R.J.}}:
\batitle{A tutorial on {B}ayesian networks for psychopathology researchers}.
\bjtitle{Psychological Methods}
\bvolume{28}(\bissue{4}),
\bfpage{947}
(\byear{2023})
\end{barticle}
\endbibitem

\bibitem[\protect\citeauthoryear{Briganti et~al.}{2024}]{briganti2024using}
\begin{barticle}
\bauthor{\bsnm{Briganti}, \binits{G.}},
\bauthor{\bsnm{Decety}, \binits{J.}},
\bauthor{\bsnm{Scutari}, \binits{M.}},
\bauthor{\bsnm{McNally}, \binits{R.J.}},
\bauthor{\bsnm{Linkowski}, \binits{P.}}:
\batitle{Using {B}ayesian networks to investigate psychological constructs: The case of empathy}.
\bjtitle{Psychological Reports}
\bvolume{127}(\bissue{5}),
\bfpage{2334}--\blpage{2346}
(\byear{2024})
\end{barticle}
\endbibitem

\bibitem[\protect\citeauthoryear{Fuster-Parra et~al.}{2014}]{fuster2014bayesian}
\begin{barticle}
\bauthor{\bsnm{Fuster-Parra}, \binits{P.}},
\bauthor{\bsnm{Garc{\'\i}a-Mas}, \binits{A.}},
\bauthor{\bsnm{Ponseti}, \binits{F.}},
\bauthor{\bsnm{Palou}, \binits{P.}},
\bauthor{\bsnm{Cruz}, \binits{J.}}:
\batitle{A {B}ayesian network to discover relationships between negative features in sport: a case study of teen players}.
\bjtitle{Quality \& Quantity}
\bvolume{48}(\bissue{3}),
\bfpage{1473}--\blpage{1491}
(\byear{2014})
\end{barticle}
\endbibitem

\bibitem[\protect\citeauthoryear{Fuster-Parra et~al.}{2015}]{fuster2015team}
\begin{barticle}
\bauthor{\bsnm{Fuster-Parra}, \binits{P.}},
\bauthor{\bsnm{Garc{\'\i}a-Mas}, \binits{A.}},
\bauthor{\bsnm{Ponseti}, \binits{F.}},
\bauthor{\bsnm{Leo}, \binits{F.M.}}:
\batitle{Team performance and collective efficacy in the dynamic psychology of competitive team: A {B}ayesian network analysis}.
\bjtitle{Human Movement Science}
\bvolume{40},
\bfpage{98}--\blpage{118}
(\byear{2015})
\end{barticle}
\endbibitem

\bibitem[\protect\citeauthoryear{Ponseti et~al.}{2019}]{ponseti2019self}
\begin{barticle}
\bauthor{\bsnm{Ponseti}, \binits{F.J.}},
\bauthor{\bsnm{Almeida}, \binits{P.L.}},
\bauthor{\bsnm{Lameiras}, \binits{J.}},
\bauthor{\bsnm{Martins}, \binits{B.}},
\bauthor{\bsnm{Olmedilla}, \binits{A.}},
\bauthor{\bsnm{L{\'o}pez-Walle}, \binits{J.}},
\bauthor{\bsnm{Reyes}, \binits{O.}},
\bauthor{\bsnm{Garcia-Mas}, \binits{A.}}:
\batitle{Self-determined motivation and competitive anxiety in athletes/students: A probabilistic study using {B}ayesian networks}.
\bjtitle{Frontiers in Psychology}
\bvolume{10},
\bfpage{1947}
(\byear{2019})
\end{barticle}
\endbibitem

\bibitem[\protect\citeauthoryear{Constantinou}{2019}]{constantinou2019dolores}
\begin{barticle}
\bauthor{\bsnm{Constantinou}, \binits{A.C.}}:
\batitle{Dolores: a model that predicts football match outcomes from all over the world}.
\bjtitle{Machine Learning}
\bvolume{108}(\bissue{1}),
\bfpage{49}--\blpage{75}
(\byear{2019})
\end{barticle}
\endbibitem

\bibitem[\protect\citeauthoryear{Constantinou et~al.}{2012}]{constantinou2012pi}
\begin{barticle}
\bauthor{\bsnm{Constantinou}, \binits{A.C.}},
\bauthor{\bsnm{Fenton}, \binits{N.E.}},
\bauthor{\bsnm{Neil}, \binits{M.}}:
\batitle{pi-football: A {B}ayesian network model for forecasting {A}ssociation {F}ootball match outcomes}.
\bjtitle{Knowledge-Based Systems}
\bvolume{36},
\bfpage{322}--\blpage{339}
(\byear{2012})
\end{barticle}
\endbibitem

\bibitem[\protect\citeauthoryear{D’Urso et~al.}{2023}]{d2023bayesian}
\begin{barticle}
\bauthor{\bsnm{D’Urso}, \binits{P.}},
\bauthor{\bsnm{De~Giovanni}, \binits{L.}},
\bauthor{\bsnm{Vitale}, \binits{V.}}:
\batitle{A {B}ayesian network to analyse basketball players’ performances: A multivariate copula-based approach}.
\bjtitle{Annals of Operations Research}
\bvolume{325}(\bissue{1}),
\bfpage{419}--\blpage{440}
(\byear{2023})
\end{barticle}
\endbibitem

\bibitem[\protect\citeauthoryear{Yung et~al.}{2025}]{yung2025using}
\begin{barticle}
\bauthor{\bsnm{Yung}, \binits{K.K.}},
\bauthor{\bsnm{Wu}, \binits{P.P.}},
\bauthor{\bsnm{F{\"u}nten}, \binits{K.}},
\bauthor{\bsnm{Hecksteden}, \binits{A.}},
\bauthor{\bsnm{Meyer}, \binits{T.}}:
\batitle{Using a {B}ayesian network to classify time to return to sport based on football injury epidemiological data}.
\bjtitle{PloS One}
\bvolume{20}(\bissue{3}),
\bfpage{0314184}
(\byear{2025})
\end{barticle}
\endbibitem

\bibitem[\protect\citeauthoryear{Zhou and Liu}{2024}]{zhou2024probability}
\begin{barticle}
\bauthor{\bsnm{Zhou}, \binits{J.Q.}},
\bauthor{\bsnm{Liu}, \binits{Y.}}:
\batitle{Probability prediction of groundstroke stances among male professional tennis players using a tree-augmented {B}ayesian network}.
\bjtitle{International Journal of Performance Analysis in Sport}
\bvolume{24}(\bissue{5}),
\bfpage{403}--\blpage{415}
(\byear{2024})
\end{barticle}
\endbibitem

\bibitem[\protect\citeauthoryear{Beuzen et~al.}{2018}]{beuzen2018comparison}
\begin{barticle}
\bauthor{\bsnm{Beuzen}, \binits{T.}},
\bauthor{\bsnm{Marshall}, \binits{L.}},
\bauthor{\bsnm{Splinter}, \binits{K.D.}}:
\batitle{A comparison of methods for discretizing continuous variables in {B}ayesian networks}.
\bjtitle{Environmental Modelling \& Software}
\bvolume{108},
\bfpage{61}--\blpage{66}
(\byear{2018})
\end{barticle}
\endbibitem

\bibitem[\protect\citeauthoryear{Nojavan et~al.}{2017}]{nojavan2017comparative}
\begin{barticle}
\bauthor{\bsnm{Nojavan}, \binits{F.}},
\bauthor{\bsnm{Qian}, \binits{S.S.}},
\bauthor{\bsnm{Stow}, \binits{C.A.}}:
\batitle{Comparative analysis of discretization methods in {B}ayesian networks}.
\bjtitle{Environmental Modelling \& Software}
\bvolume{87},
\bfpage{64}--\blpage{71}
(\byear{2017})
\end{barticle}
\endbibitem

\bibitem[\protect\citeauthoryear{Lauritzen and Wermuth}{1989}]{lauritzen1989graphical}
\begin{botherref}
\oauthor{\bsnm{Lauritzen}, \binits{S.L.}},
\oauthor{\bsnm{Wermuth}, \binits{N.}}:
Graphical models for associations between variables, some of which are qualitative and some quantitative.
The Annals of Statistics,
31--57
(1989)
\end{botherref}
\endbibitem

\bibitem[\protect\citeauthoryear{Hoff}{2007}]{hoff2007extending}
\begin{barticle}
\bauthor{\bsnm{Hoff}, \binits{P.}}:
\batitle{Extending the rank likelihood for semiparametric copula estimation}.
\bjtitle{Annals of Applied Statistics}
\bvolume{1},
\bfpage{265}--\blpage{283}
(\byear{2007})
\end{barticle}
\endbibitem

\bibitem[\protect\citeauthoryear{Cui et~al.}{2016}]{cui2016copula}
\begin{bchapter}
\bauthor{\bsnm{Cui}, \binits{R.}},
\bauthor{\bsnm{Groot}, \binits{P.}},
\bauthor{\bsnm{Heskes}, \binits{T.}}:
\bctitle{Copula {PC} algorithm for causal discovery from mixed data}.
In: \bbtitle{Joint European Conference on Machine Learning and Knowledge Discovery in Databases},
pp. \bfpage{377}--\blpage{392}
(\byear{2016}).
\bcomment{Springer}
\end{bchapter}
\endbibitem

\bibitem[\protect\citeauthoryear{Cai et~al.}{2022}]{cai2022causal}
\begin{barticle}
\bauthor{\bsnm{Cai}, \binits{Z.}},
\bauthor{\bsnm{Xi}, \binits{D.}},
\bauthor{\bsnm{Zhu}, \binits{X.}},
\bauthor{\bsnm{Li}, \binits{R.}}:
\batitle{Causal discoveries for high dimensional mixed data}.
\bjtitle{Statistics in Medicine}
\bvolume{41}(\bissue{24}),
\bfpage{4924}--\blpage{4940}
(\byear{2022})
\end{barticle}
\endbibitem

\bibitem[\protect\citeauthoryear{Castelletti}{2024}]{castelletti2024learning}
\begin{barticle}
\bauthor{\bsnm{Castelletti}, \binits{F.}}:
\batitle{Learning {B}ayesian networks: A copula approach for mixed-type data}.
\bjtitle{Psychometrika}
\bvolume{89}(\bissue{2}),
\bfpage{658}--\blpage{686}
(\byear{2024})
\end{barticle}
\endbibitem

\bibitem[\protect\citeauthoryear{Tsamardinos et~al.}{2006}]{tsamardinos2006max}
\begin{barticle}
\bauthor{\bsnm{Tsamardinos}, \binits{I.}},
\bauthor{\bsnm{Brown}, \binits{L.E.}},
\bauthor{\bsnm{Aliferis}, \binits{C.F.}}:
\batitle{The max-min hill-climbing {B}ayesian network structure learning algorithm}.
\bjtitle{Machine Learning}
\bvolume{65}(\bissue{1}),
\bfpage{31}--\blpage{78}
(\byear{2006})
\end{barticle}
\endbibitem

\bibitem[\protect\citeauthoryear{Scutari and Nagarajan}{2013}]{scutari2013identifying}
\begin{barticle}
\bauthor{\bsnm{Scutari}, \binits{M.}},
\bauthor{\bsnm{Nagarajan}, \binits{R.}}:
\batitle{Identifying significant edges in graphical models of molecular networks}.
\bjtitle{Artificial Intelligence in Medicine}
\bvolume{57}(\bissue{3}),
\bfpage{207}--\blpage{217}
(\byear{2013})
\end{barticle}
\endbibitem

\bibitem[\protect\citeauthoryear{Caravagna and Ramazzotti}{2021}]{caravagna2021learning}
\begin{barticle}
\bauthor{\bsnm{Caravagna}, \binits{G.}},
\bauthor{\bsnm{Ramazzotti}, \binits{D.}}:
\batitle{Learning the structure of {B}ayesian networks via the bootstrap}.
\bjtitle{Neurocomputing}
\bvolume{448},
\bfpage{48}--\blpage{59}
(\byear{2021})
\end{barticle}
\endbibitem

\bibitem[\protect\citeauthoryear{Pearl}{2009}]{pearl2009causality}
\begin{bbook}
\bauthor{\bsnm{Pearl}, \binits{J.}}:
\bbtitle{Causality}.
\bpublisher{Cambridge University Press},
\blocation{Cambridge}
(\byear{2009})
\end{bbook}
\endbibitem

\bibitem[\protect\citeauthoryear{Verma and Pearl}{1990}]{verma1990equivalence}
\begin{bchapter}
\bauthor{\bsnm{Verma}, \binits{T.}},
\bauthor{\bsnm{Pearl}, \binits{J.}}:
\bctitle{Equivalence and synthesis of causal models}.
In: \bbtitle{Proceedings of the 6th Conference on Uncertainty in Artificial Intelligence},
pp. \bfpage{220}--\blpage{227}
(\byear{1990})
\end{bchapter}
\endbibitem

\bibitem[\protect\citeauthoryear{Andersson et~al.}{1997}]{andersson1997characterization}
\begin{barticle}
\bauthor{\bsnm{Andersson}, \binits{S.A.}},
\bauthor{\bsnm{Madigan}, \binits{D.}},
\bauthor{\bsnm{Perlman}, \binits{M.D.}}:
\batitle{A characterization of {M}arkov equivalence classes for acyclic digraphs}.
\bjtitle{The Annals of Statistics}
\bvolume{25}(\bissue{2}),
\bfpage{505}--\blpage{541}
(\byear{1997})
\end{barticle}
\endbibitem

\bibitem[\protect\citeauthoryear{Kitson et~al.}{2023}]{kitson2023survey}
\begin{barticle}
\bauthor{\bsnm{Kitson}, \binits{N.K.}},
\bauthor{\bsnm{Constantinou}, \binits{A.C.}},
\bauthor{\bsnm{Guo}, \binits{Z.}},
\bauthor{\bsnm{Liu}, \binits{Y.}},
\bauthor{\bsnm{Chobtham}, \binits{K.}}:
\batitle{A survey of {B}ayesian network structure learning}.
\bjtitle{Artificial Intelligence Review}
\bvolume{56}(\bissue{8}),
\bfpage{8721}--\blpage{8814}
(\byear{2023})
\end{barticle}
\endbibitem

\bibitem[\protect\citeauthoryear{Scanagatta et~al.}{2019}]{scanagatta2019survey}
\begin{barticle}
\bauthor{\bsnm{Scanagatta}, \binits{M.}},
\bauthor{\bsnm{Salmer{\'o}n}, \binits{A.}},
\bauthor{\bsnm{Stella}, \binits{F.}}:
\batitle{A survey on {B}ayesian network structure learning from data}.
\bjtitle{Progress in Artificial Intelligence}
\bvolume{8}(\bissue{4}),
\bfpage{425}--\blpage{439}
(\byear{2019})
\end{barticle}
\endbibitem

\bibitem[\protect\citeauthoryear{Spirtes et~al.}{2000}]{spirtes2000causation}
\begin{bbook}
\bauthor{\bsnm{Spirtes}, \binits{P.}},
\bauthor{\bsnm{Glymour}, \binits{C.N.}},
\bauthor{\bsnm{Scheines}, \binits{R.}}:
\bbtitle{Causation, Prediction, and Search}.
\bpublisher{MIT press},
\blocation{London}
(\byear{2000})
\end{bbook}
\endbibitem

\bibitem[\protect\citeauthoryear{Colombo and Maathuis}{2014}]{colombo2014order}
\begin{barticle}
\bauthor{\bsnm{Colombo}, \binits{D.}},
\bauthor{\bsnm{Maathuis}, \binits{M.H.}}:
\batitle{Order-independent constraint-based causal structure learning}.
\bjtitle{Journal of Machine Learning Research}
\bvolume{15}(\bissue{1}),
\bfpage{3741}--\blpage{3782}
(\byear{2014})
\end{barticle}
\endbibitem

\bibitem[\protect\citeauthoryear{Schwarz}{1978}]{Schwarz1978}
\begin{barticle}
\bauthor{\bsnm{Schwarz}, \binits{G.E.}}:
\batitle{Estimating the dimension of a model}.
\bjtitle{Annals of Statistics}
\bvolume{6}(\bissue{2}),
\bfpage{461}--\blpage{464}
(\byear{1978})
\end{barticle}
\endbibitem

\bibitem[\protect\citeauthoryear{Scutari}{2010}]{scutari2010learning}
\begin{barticle}
\bauthor{\bsnm{Scutari}, \binits{M.}}:
\batitle{Learning {B}ayesian networks with the bnlearn {R} package}.
\bjtitle{Journal of Statistical Software}
\bvolume{35},
\bfpage{1}--\blpage{22}
(\byear{2010})
\end{barticle}
\endbibitem

\bibitem[\protect\citeauthoryear{Constantinou et~al.}{2021}]{constantinou2021large}
\begin{barticle}
\bauthor{\bsnm{Constantinou}, \binits{A.C.}},
\bauthor{\bsnm{Liu}, \binits{Y.}},
\bauthor{\bsnm{Chobtham}, \binits{K.}},
\bauthor{\bsnm{Guo}, \binits{Z.}},
\bauthor{\bsnm{Kitson}, \binits{N.K.}}:
\batitle{Large-scale empirical validation of {B}ayesian network structure learning algorithms with noisy data}.
\bjtitle{International Journal of Approximate Reasoning}
\bvolume{131},
\bfpage{151}--\blpage{188}
(\byear{2021})
\end{barticle}
\endbibitem

\bibitem[\protect\citeauthoryear{Scutari et~al.}{2019}]{scutari2019learns}
\begin{barticle}
\bauthor{\bsnm{Scutari}, \binits{M.}},
\bauthor{\bsnm{Graafland}, \binits{C.E.}},
\bauthor{\bsnm{Guti{\'e}rrez}, \binits{J.M.}}:
\batitle{Who learns better {B}ayesian network structures: Accuracy and speed of structure learning algorithms}.
\bjtitle{International Journal of Approximate Reasoning}
\bvolume{115},
\bfpage{235}--\blpage{253}
(\byear{2019})
\end{barticle}
\endbibitem

\bibitem[\protect\citeauthoryear{Geiger and Heckerman}{1994}]{geiger1994learning}
\begin{bchapter}
\bauthor{\bsnm{Geiger}, \binits{D.}},
\bauthor{\bsnm{Heckerman}, \binits{D.}}:
\bctitle{Learning {G}aussian networks}.
In: \bbtitle{Proceedings of the 10th Conference on Uncertainty in Artificial Intelligence},
pp. \bfpage{235}--\blpage{243}
(\byear{1994})
\end{bchapter}
\endbibitem

\bibitem[\protect\citeauthoryear{Richardson and Spirtes}{2002}]{richardson2002ancestral}
\begin{barticle}
\bauthor{\bsnm{Richardson}, \binits{T.}},
\bauthor{\bsnm{Spirtes}, \binits{P.}}:
\batitle{Ancestral graph {M}arkov models}.
\bjtitle{The Annals of Statistics}
\bvolume{30}(\bissue{4}),
\bfpage{962}--\blpage{1030}
(\byear{2002})
\end{barticle}
\endbibitem

\bibitem[\protect\citeauthoryear{Murphy}{2012}]{murphy2012machine}
\begin{bbook}
\bauthor{\bsnm{Murphy}, \binits{K.P.}}:
\bbtitle{Machine Learning: A Probabilistic Perspective}.
\bpublisher{MIT Press},
\blocation{London}
(\byear{2012})
\end{bbook}
\endbibitem

\bibitem[\protect\citeauthoryear{Kalisch and B{\"u}hlman}{2007}]{kalisch2007estimating}
\begin{botherref}
\oauthor{\bsnm{Kalisch}, \binits{M.}},
\oauthor{\bsnm{B{\"u}hlman}, \binits{P.}}:
Estimating high-dimensional directed acyclic graphs with the {PC-algorithm}.
Journal of Machine Learning Research
\textbf{8}(3)
(2007)
\end{botherref}
\endbibitem

\bibitem[\protect\citeauthoryear{Harris and Drton}{2013}]{harris2013pc}
\begin{barticle}
\bauthor{\bsnm{Harris}, \binits{N.}},
\bauthor{\bsnm{Drton}, \binits{M.}}:
\batitle{{PC} algorithm for nonparanormal graphical models}.
\bjtitle{The Journal of Machine Learning Research}
\bvolume{14}(\bissue{1}),
\bfpage{3365}--\blpage{3383}
(\byear{2013})
\end{barticle}
\endbibitem

\bibitem[\protect\citeauthoryear{Andrews et~al.}{2018}]{andrews2018scoring}
\begin{barticle}
\bauthor{\bsnm{Andrews}, \binits{B.}},
\bauthor{\bsnm{Ramsey}, \binits{J.}},
\bauthor{\bsnm{Cooper}, \binits{G.F.}}:
\batitle{Scoring {B}ayesian networks of mixed variables}.
\bjtitle{International Journal of Data Science and Analytics}
\bvolume{6}(\bissue{1}),
\bfpage{3}--\blpage{18}
(\byear{2018})
\end{barticle}
\endbibitem

\bibitem[\protect\citeauthoryear{Raghu et~al.}{2018}]{raghu2018evaluation}
\begin{bchapter}
\bauthor{\bsnm{Raghu}, \binits{V.K.}},
\bauthor{\bsnm{Poon}, \binits{A.}},
\bauthor{\bsnm{Benos}, \binits{P.V.}}:
\bctitle{Evaluation of causal structure learning methods on mixed data types}.
In: \bbtitle{Proceedings of 2018 ACM SIGKDD Workshop on Causal Discovery},
pp. \bfpage{48}--\blpage{65}
(\byear{2018}).
\bcomment{PMLR}
\end{bchapter}
\endbibitem

\bibitem[\protect\citeauthoryear{Talvitie et~al.}{2019}]{talvitie2019learning}
\begin{barticle}
\bauthor{\bsnm{Talvitie}, \binits{T.}},
\bauthor{\bsnm{Eggeling}, \binits{R.}},
\bauthor{\bsnm{Koivisto}, \binits{M.}}:
\batitle{Learning {B}ayesian networks with local structure, mixed variables, and exact algorithms}.
\bjtitle{International Journal of Approximate Reasoning}
\bvolume{115},
\bfpage{69}--\blpage{95}
(\byear{2019})
\end{barticle}
\endbibitem

\bibitem[\protect\citeauthoryear{Tsagris et~al.}{2018}]{tsagris2018constraint}
\begin{barticle}
\bauthor{\bsnm{Tsagris}, \binits{M.}},
\bauthor{\bsnm{Borboudakis}, \binits{G.}},
\bauthor{\bsnm{Lagani}, \binits{V.}},
\bauthor{\bsnm{Tsamardinos}, \binits{I.}}:
\batitle{Constraint-based causal discovery with mixed data}.
\bjtitle{International Journal of Data Science and Analytics}
\bvolume{6}(\bissue{1}),
\bfpage{19}--\blpage{30}
(\byear{2018})
\end{barticle}
\endbibitem

\bibitem[\protect\citeauthoryear{Liu et~al.}{2009}]{liu2009nonparanormal}
\begin{botherref}
\oauthor{\bsnm{Liu}, \binits{H.}},
\oauthor{\bsnm{Lafferty}, \binits{J.}},
\oauthor{\bsnm{Wasserman}, \binits{L.}}:
The nonparanormal: Semiparametric estimation of high dimensional undirected graphs.
Journal of Machine Learning Research
\textbf{10}(10)
(2009)
\end{botherref}
\endbibitem

\bibitem[\protect\citeauthoryear{Dobra and Lenkoski}{2011}]{dobra2011copula}
\begin{barticle}
\bauthor{\bsnm{Dobra}, \binits{A.}},
\bauthor{\bsnm{Lenkoski}, \binits{A.}}:
\batitle{Copula {G}aussian graphical models and their application to modeling functional disability data}.
\bjtitle{The Annals of Applied Statistics}
\bvolume{2A},
\bfpage{969}--\blpage{993}
(\byear{2011})
\end{barticle}
\endbibitem

\bibitem[\protect\citeauthoryear{Kalisch et~al.}{2012}]{kalisch2012causal}
\begin{barticle}
\bauthor{\bsnm{Kalisch}, \binits{M.}},
\bauthor{\bsnm{M{\"a}chler}, \binits{M.}},
\bauthor{\bsnm{Colombo}, \binits{D.}},
\bauthor{\bsnm{Maathuis}, \binits{M.H.}},
\bauthor{\bsnm{B{\"u}hlmann}, \binits{P.}}:
\batitle{Causal inference using graphical models with the {R} package pcalg}.
\bjtitle{Journal of Statistical Software}
\bvolume{47},
\bfpage{1}--\blpage{26}
(\byear{2012})
\end{barticle}
\endbibitem

\bibitem[\protect\citeauthoryear{Leonelli et~al.}{2023}]{leonelli2023sensitivity}
\begin{barticle}
\bauthor{\bsnm{Leonelli}, \binits{M.}},
\bauthor{\bsnm{Ramanathan}, \binits{R.}},
\bauthor{\bsnm{Wilkerson}, \binits{R.L.}}:
\batitle{Sensitivity and robustness analysis in {B}ayesian networks with the bnmonitor {R} package}.
\bjtitle{Knowledge-Based Systems}
\bvolume{278},
\bfpage{110882}
(\byear{2023})
\end{barticle}
\endbibitem

\bibitem[\protect\citeauthoryear{McCrae and Costa}{2008}]{mccrae2008fivefactor}
\begin{bchapter}
\bauthor{\bsnm{McCrae}, \binits{R.R.}},
\bauthor{\bsnm{Costa}, \binits{P.T.}}:
\bctitle{The five-factor theory of personality}.
In: \beditor{\bsnm{John}, \binits{O.P.}},
\beditor{\bsnm{Robins}, \binits{R.W.}},
\beditor{\bsnm{Pervin}, \binits{L.A.}} (eds.)
\bbtitle{Handbook of Personality: Theory and Research},
pp. \bfpage{159}--\blpage{181}.
\bpublisher{Guilford Press},
\blocation{New York, NY}
(\byear{2008})
\end{bchapter}
\endbibitem

\bibitem[\protect\citeauthoryear{Robazza et~al.}{2009}]{robazza2009ipps}
\begin{barticle}
\bauthor{\bsnm{Robazza}, \binits{C.}},
\bauthor{\bsnm{Bortoli}, \binits{L.}},
\bauthor{\bsnm{Gramaccioni}, \binits{G.}}:
\batitle{L'inventario psicologico della prestazione sportiva (ipps-48)}.
\bjtitle{Giornale Italiano di Psicologia dello Sport}
\bvolume{4},
\bfpage{14}--\blpage{20}
(\byear{2009})
\end{barticle}
\endbibitem

\bibitem[\protect\citeauthoryear{Rosenberg}{1965}]{rosenberg1965selfimage}
\begin{bbook}
\bauthor{\bsnm{Rosenberg}, \binits{M.}}:
\bbtitle{Society and the Adolescent Self-image}.
\bpublisher{Princeton University Press},
\blocation{Princeton, NJ}
(\byear{1965})
\end{bbook}
\endbibitem

\bibitem[\protect\citeauthoryear{Leonelli}{2025}]{leonelli2025bnrep}
\begin{barticle}
\bauthor{\bsnm{Leonelli}, \binits{M.}}:
\batitle{bn{R}ep: A repository of {B}ayesian networks from the academic literature}.
\bjtitle{Neurocomputing}
\bvolume{624},
\bfpage{129502}
(\byear{2025})
\end{barticle}
\endbibitem

\bibitem[\protect\citeauthoryear{Ballester-Ripoll and Leonelli}{2022}]{ballester2022computing}
\begin{barticle}
\bauthor{\bsnm{Ballester-Ripoll}, \binits{R.}},
\bauthor{\bsnm{Leonelli}, \binits{M.}}:
\batitle{Computing {S}obol indices in probabilistic graphical models}.
\bjtitle{Reliability Engineering \& System Safety}
\bvolume{225},
\bfpage{108573}
(\byear{2022})
\end{barticle}
\endbibitem

\end{thebibliography}

\newpage 
\appendix

\section{Additional Conditional Probability Inferences}
\begin{table*}[h]
\centering
\footnotesize
\caption{Conditional distributions of \emph{Worry} (columns; categories One–Six) given its main precursors (rows), compared to the baseline distribution.}
\label{tab:worry_shift}
\setlength{\tabcolsep}{5.5pt}
\begin{tabular}{lrrrrrr}
\toprule
 & One & Two & Three & Four & Five & Six \\
\midrule
\multicolumn{7}{l}{\textbf{Baseline (no evidence)}}\\
Baseline & 0.07 & 0.16 & 0.29 & 0.34 & 0.10 & 0.04 \\
\addlinespace[4pt]

\multicolumn{7}{l}{\textbf{Given Neuroticism}}\\
One   & 0.12 & 0.30 & 0.23 & 0.12 & 0.12 & 0.12 \\
Two   & 0.07 & 0.28 & 0.25 & 0.34 & 0.03 & 0.03 \\
Three & 0.07 & 0.12 & 0.37 & 0.36 & 0.07 & 0.02 \\
Four  & 0.07 & 0.12 & 0.23 & 0.35 & 0.18 & 0.05 \\
Five  & 0.13 & 0.13 & 0.19 & 0.18 & 0.25 & 0.13 \\
\addlinespace[4pt]

\multicolumn{7}{l}{\textbf{Given Concentration disruption}}\\
One   & 0.23 & 0.14 & 0.35 & 0.15 & 0.07 & 0.07 \\
Two   & 0.09 & 0.29 & 0.25 & 0.29 & 0.06 & 0.02 \\
Three & 0.02 & 0.07 & 0.26 & 0.52 & 0.10 & 0.02 \\
Four  & 0.04 & 0.07 & 0.45 & 0.20 & 0.18 & 0.06 \\
Five  & 0.15 & 0.15 & 0.15 & 0.15 & 0.25 & 0.15 \\
Six   & 0.17 & 0.17 & 0.17 & 0.17 & 0.17 & 0.17 \\
\bottomrule
\end{tabular}
\end{table*}

\begin{table*}[t]
\centering
\footnotesize
\caption{Conditional distributions of \emph{Self-confidence} (columns; categories One–Six) given its main precursors (rows), compared to the baseline distribution.}
\label{tab:selfconf_shift}
\setlength{\tabcolsep}{5pt}
\begin{tabular}{lrrrrrr}
\toprule
 & One & Two & Three & Four & Five & Six \\
\midrule
\multicolumn{7}{l}{\textbf{Baseline (no evidence)}}\\
Baseline & 0.02 & 0.13 & 0.22 & 0.38 & 0.14 & 0.10 \\
\addlinespace[4pt]

\multicolumn{7}{l}{\textbf{Given Extraversion}}\\
One   & 0.08 & 0.34 & 0.30 & 0.23 & 0.01 & 0.05 \\
Two   & 0.05 & 0.22 & 0.29 & 0.34 & 0.07 & 0.04 \\
Three & 0.01 & 0.14 & 0.23 & 0.40 & 0.13 & 0.08 \\
Four  & 0.01 & 0.08 & 0.18 & 0.43 & 0.19 & 0.10 \\
Five  & 0.00 & 0.07 & 0.15 & 0.34 & 0.19 & 0.24 \\
\addlinespace[4pt]

\multicolumn{7}{l}{\textbf{Given Self-talk}}\\
One   & 0.06 & 0.26 & 0.33 & 0.22 & 0.10 & 0.03 \\
Two   & 0.03 & 0.18 & 0.28 & 0.33 & 0.14 & 0.04 \\
Three & 0.02 & 0.14 & 0.22 & 0.41 & 0.14 & 0.08 \\
Four  & 0.02 & 0.12 & 0.20 & 0.41 & 0.14 & 0.11 \\
Five  & 0.01 & 0.10 & 0.17 & 0.38 & 0.15 & 0.19 \\
Six   & 0.00 & 0.08 & 0.14 & 0.33 & 0.16 & 0.30 \\
\addlinespace[4pt]

\multicolumn{7}{l}{\textbf{Given Goal setting}}\\
One   & 0.33 & 0.65 & 0.00 & 0.00 & 0.00 & 0.00 \\
Two   & 0.03 & 0.32 & 0.45 & 0.16 & 0.03 & 0.00 \\
Three & 0.00 & 0.12 & 0.30 & 0.44 & 0.14 & 0.00 \\
Four  & 0.00 & 0.05 & 0.11 & 0.63 & 0.13 & 0.07 \\
Five  & 0.00 & 0.00 & 0.16 & 0.11 & 0.42 & 0.31 \\
Six   & 0.00 & 0.00 & 0.00 & 0.20 & 0.10 & 0.69 \\
\addlinespace[4pt]

\multicolumn{7}{l}{\textbf{Given Emotional arousal}}\\
One   & 0.32 & 0.32 & 0.32 & 0.01 & 0.01 & 0.01 \\
Two   & 0.08 & 0.35 & 0.31 & 0.23 & 0.00 & 0.04 \\
Three & 0.00 & 0.18 & 0.32 & 0.39 & 0.09 & 0.04 \\
Four  & 0.00 & 0.04 & 0.13 & 0.58 & 0.19 & 0.06 \\
Five  & 0.00 & 0.00 & 0.13 & 0.19 & 0.37 & 0.31 \\
Six   & 0.00 & 0.00 & 0.00 & 0.11 & 0.22 & 0.66 \\
\addlinespace[4pt]

\multicolumn{7}{l}{\textbf{Given Mental practice}}\\
One   & 0.00 & 0.33 & 0.22 & 0.11 & 0.33 & 0.00 \\
Two   & 0.04 & 0.25 & 0.38 & 0.17 & 0.13 & 0.02 \\
Three & 0.02 & 0.07 & 0.22 & 0.53 & 0.11 & 0.05 \\
Four  & 0.00 & 0.05 & 0.05 & 0.63 & 0.13 & 0.13 \\
Five  & 0.00 & 0.00 & 0.00 & 0.00 & 0.28 & 0.70 \\
Six   & 0.01 & 0.01 & 0.01 & 0.01 & 0.01 & 0.96 \\
\addlinespace[4pt]

\multicolumn{7}{l}{\textbf{Given Match preparation}}\\
One   & 0.12 & 0.40 & 0.30 & 0.13 & 0.04 & 0.01 \\
Two   & 0.03 & 0.19 & 0.35 & 0.26 & 0.13 & 0.03 \\
Three & 0.02 & 0.15 & 0.22 & 0.40 & 0.16 & 0.04 \\
Four  & 0.01 & 0.09 & 0.17 & 0.51 & 0.13 & 0.09 \\
Five  & 0.00 & 0.06 & 0.13 & 0.35 & 0.17 & 0.29 \\
Six   & 0.00 & 0.06 & 0.11 & 0.14 & 0.18 & 0.51 \\
\addlinespace[4pt]

\multicolumn{7}{l}{\textbf{Given Self-esteem}}\\
One   & 0.09 & 0.02 & 0.21 & 0.16 & 0.31 & 0.20 \\
Two   & 0.01 & 0.22 & 0.30 & 0.41 & 0.02 & 0.04 \\
Three & 0.03 & 0.16 & 0.23 & 0.41 & 0.12 & 0.05 \\
Four  & 0.00 & 0.00 & 0.12 & 0.30 & 0.28 & 0.30 \\
\bottomrule
\end{tabular}
\end{table*}

\begin{table*}[t]
\centering
\footnotesize
\caption{Conditional distributions of \emph{Self-esteem} (columns; categories One–Four) given its main precursors (rows), compared to the baseline distribution.}
\label{tab:selfesteem_shift}
\setlength{\tabcolsep}{5pt}
\begin{tabular}{lrrrr}
\toprule
 & One & Two & Three & Four \\
\midrule
\multicolumn{5}{l}{\textbf{Baseline (no evidence)}}\\
Baseline & 0.01 & 0.19 & 0.57 & 0.22 \\
\addlinespace[4pt]

\multicolumn{5}{l}{\textbf{Given Extraversion}}\\
One   & 0.15 & 0.55 & 0.15 & 0.15 \\
Two   & 0.03 & 0.36 & 0.59 & 0.03 \\
Three & 0.00 & 0.24 & 0.65 & 0.12 \\
Four  & 0.00 & 0.06 & 0.54 & 0.40 \\
Five  & 0.00 & 0.06 & 0.48 & 0.46 \\
\addlinespace[4pt]

\multicolumn{5}{l}{\textbf{Given Goal setting}}\\
One   & 0.02 & 0.27 & 0.68 & 0.03 \\
Two   & 0.01 & 0.28 & 0.63 & 0.08 \\
Three & 0.01 & 0.23 & 0.60 & 0.16 \\
Four  & 0.01 & 0.17 & 0.58 & 0.24 \\
Five  & 0.02 & 0.09 & 0.45 & 0.44 \\
Six   & 0.02 & 0.09 & 0.36 & 0.53 \\
\addlinespace[4pt]

\multicolumn{5}{l}{\textbf{Given Self-confidence}}\\
One   & 0.05 & 0.15 & 0.75 & 0.05 \\
Two   & 0.00 & 0.32 & 0.68 & 0.00 \\
Three & 0.01 & 0.27 & 0.60 & 0.12 \\
Four  & 0.00 & 0.21 & 0.61 & 0.17 \\
Five  & 0.02 & 0.02 & 0.51 & 0.45 \\
Six   & 0.02 & 0.07 & 0.27 & 0.64 \\
\addlinespace[4pt]

\multicolumn{5}{l}{\textbf{Given Emotional arousal}}\\
One   & 0.03 & 0.24 & 0.65 & 0.07 \\
Two   & 0.03 & 0.33 & 0.58 & 0.06 \\
Three & 0.01 & 0.23 & 0.63 & 0.13 \\
Four  & 0.01 & 0.15 & 0.58 & 0.27 \\
Five  & 0.00 & 0.07 & 0.44 & 0.49 \\
Six   & 0.00 & 0.03 & 0.39 & 0.58 \\
\bottomrule
\end{tabular}
\end{table*}

\end{document}